\documentclass[lettersize,journal]{IEEEtran}
\usepackage{amsmath,amsfonts}
\usepackage{algorithmic}
\usepackage{algorithm}
\usepackage{array}
\usepackage[caption=false,font=normalsize,labelfont=sf,textfont=sf]{subfig}
\usepackage{textcomp}
\usepackage{stfloats}
\usepackage{url}
\usepackage{verbatim}
\usepackage{graphicx}
\usepackage{cite}
\usepackage[table]{xcolor}
\usepackage{arydshln}
\usepackage{lipsum}
\usepackage{multirow}
\usepackage{threeparttable}
\usepackage{tikz}
\usepackage{color, colortbl}
\usepackage{makecell}
\usepackage{booktabs}
\usepackage{amssymb}
\usepackage{xurl}
\usepackage{hyperref}

\definecolor{deeppeach}{rgb}{1.0, 0.8, 0.64}
\definecolor{lightgray}{rgb}{0.83, 0.83, 0.83}

\begin{document}

\title{Exploring Disentangled and Controllable Human Image Synthesis: From End-to-End \\ to Stage-by-Stage}

\author{Zhengwentai Sun, Chenghong Li, Hongjie Liao, Xihe Yang, Keru Zheng, Heyuan Li, \\Yihao Zhi, Shuliang Ning, Shuguang Cui~\IEEEmembership{Fellow,~IEEE} and Xiaoguang Han$^{\dag}$~\IEEEmembership{Member,~IEEE}
\IEEEcompsocitemizethanks{
    \IEEEcompsocthanksitem Z. Sun, C. Li, H. Liao, X. Yang, K. Zheng, H. Li, Y. Zhi, S. Ning, S. Cui and X. Han are currently with the School of Science and Engineering, The Chinese University of Hong Kong, Shenzhen. Z. Sun, C. Li, Y. Zhi, S. Cui and X. Han are also with the Future Network of Intelligence Institute, CUHK-Shenzhen. (email: \{zhengwentaisun, chenghongli, hongjieliao, xiheyang1, keruzheng, heyuanli, yihaozhi1, shuliangning\}@link.cuhk.edu.cn, \{shuguangcui, hanxiaoguang\}@cuhk.edu.cn).

    \IEEEcompsocthanksitem $\dag$ denotes corresponding author.
    
}}

\maketitle

\begin{abstract}
Achieving fine-grained controllability in human image synthesis is a long-standing challenge in computer vision. Existing methods primarily focus on either facial synthesis or near-frontal body generation, with limited ability to simultaneously control key factors such as viewpoint, pose, clothing, and identity in a disentangled manner. In this paper, we introduce a new disentangled and controllable human synthesis task, which explicitly separates and manipulates these four factors within a unified framework. We first develop an end-to-end generative model trained on MVHumanNet for factor disentanglement. However, the domain gap between MVHumanNet and in-the-wild data produces unsatisfactory results, motivating the exploration of virtual try-on (VTON) dataset as a potential solution. Through experiments, we observe that simply incorporating the VTON dataset as additional data to train the end-to-end model degrades performance, primarily due to the inconsistency in data forms between the two datasets, which disrupts the disentanglement process. To better leverage both datasets, we propose a stage-by-stage framework that decomposes human image generation into three sequential steps: clothed A-pose generation, back-view synthesis, and pose and view control. This structured pipeline enables better dataset utilization at different stages, significantly improving controllability and generalization, especially for in-the-wild scenarios. Extensive experiments demonstrate that our stage-by-stage approach outperforms end-to-end models in both visual fidelity and disentanglement quality, offering a scalable solution for real-world tasks.
\end{abstract}


\begin{figure*}[t]
    \centering
    \includegraphics[width=0.98\linewidth]{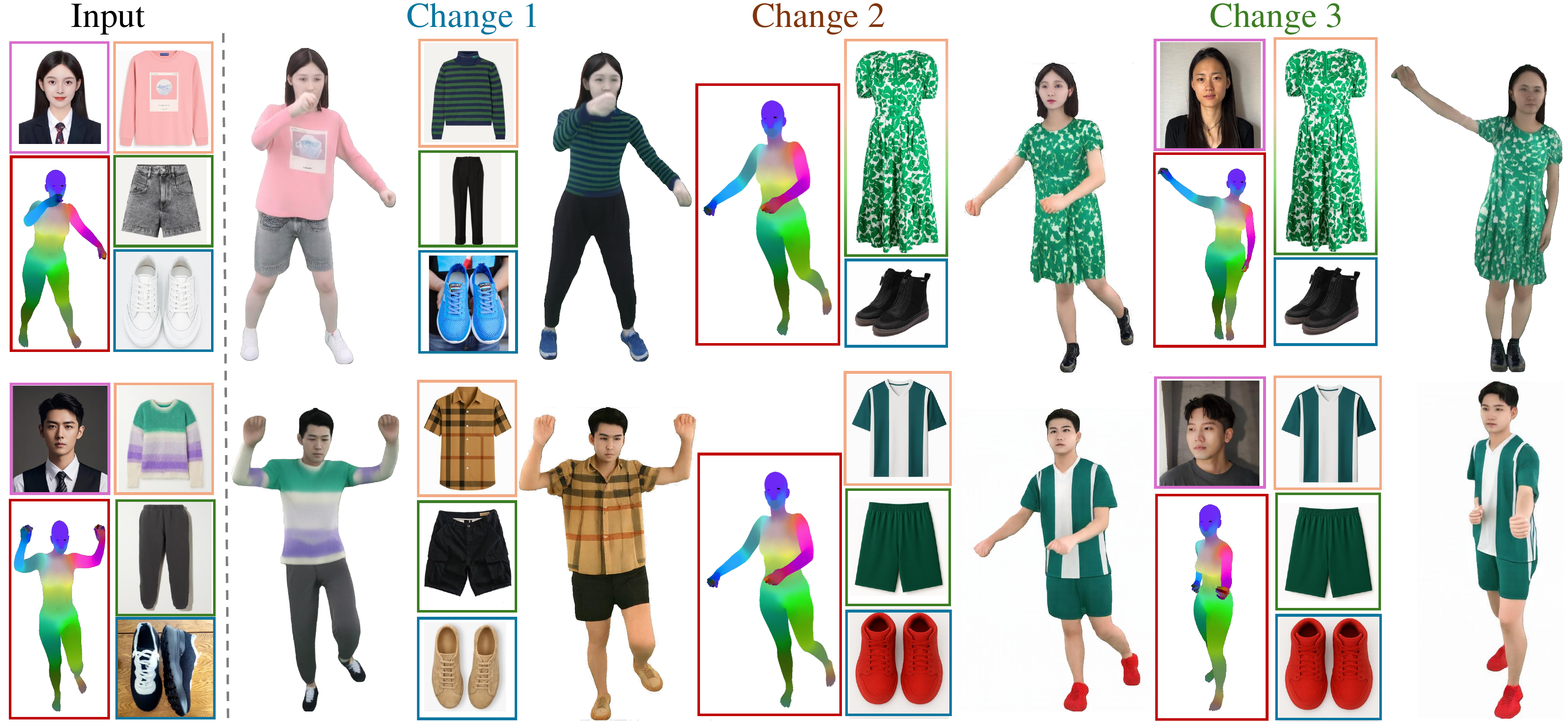}
    \caption{We propose a new task focused on explicit disentanglement of key human attributes within a unified framework, enabling fine-grained controllability in human synthesis. Moreover, we explore to train an end-to-end model and further design a stage-by-stage pipeline for generating human images with customizable inputs, offering enhanced flexibility and superior control over the synthesis process.}
    \label{fig:teaser}
\end{figure*}

\section{Introduction}
\IEEEPARstart{S}{ynthesizing} human-centric images is crucial for various applications like virtual reality, the metaverse, video games, and digital fashion. Advances in GANs \cite{goodfellow2014generative, Karras_2020_CVPR} and large-scale facial datasets \cite{liu2015faceattributes, Karras_2019_CVPR} have enabled realistic face generation \cite{karras2018progressive}, with subsequent research focusing on controlling facial appearances and expressions \cite{Choi_2018_CVPR, Lu_2018_ECCV, yang2020deep}, though often limited to frontal views. The introduction of 3D-aware GANs, EG3D \cite{Chan_2022_CVPR}, brought view controllability, paving the way for unified control over viewpoint, expression, and appearance \cite{An_2023_CVPR, li2024spherehead, xie2023high, Lin20223DGI, Bai2024Realtime3P}. Compared with human faces, full-body image generation is more complex due to clothing variations and pose deformations. 

Early GAN-based full-body image generation methods \cite{Fruhstuck_2022_CVPR,fu2022stylegan,Sarkar2021HumanGAN} supporting pose or appearance editing but restricted to near-frontal views. Following the key idea of EG3D \cite{Chan_2022_CVPR}, EVA3D \cite{hong2023evad} also tried to further enable view controllability of human image generation. There are also a specific topic in this area which focus on clothing image synthesis, i.e., virtual try-on \cite{Choi_2021_CVPR,Morelli_2022_CVPR,lee2022high,choi2025IDM}.  To achieve free-view human image synthesis, disentanglement and controllability have been key research focuses. However, a unified framework addressing all requirements remains challenging, primarily due to the lack of datasets with disentangled factors. The recent release of the multi-view human video dataset, MVHumanNet \cite{Xiong_2024_CVPR, li2025mvhumannet++}, provides the possibility to take a step forward in this research direction.

To achieve fine-grained controllability over human synthesis, we propose a task that emphasizes explicit disentanglement of facial identity, clothing, pose, and viewpoint within a unified framework. A natural approach to solving this problem is to train an end-to-end model that synthesizes human images while explicitly conditioning on these factors.
We attempted to conduct experiments on MVHumanNet to train an end-to-end model. While the model demonstrates a degree of controllability, its robustness and generalizability remain limited when trained exclusively on this dataset. These limitations primarily stem from inherent issues within the dataset itself. Specifically, MVHumanNet is collected in an indoor environment, which introduces a domain gap when compared to in-the-wild data.

Given these challenges, we explore to incorporate VTON dataset, which contains high-quality clothing images with more detailed textures, combined with MVHumanNet to train end-to-end model. However, our experiments reveal that simply incorporating VTON dataset as additional data to train end-to-end model result in performance degradation rather than improvement. We attribute this decline to the inconsistency in data forms between the two datasets when training the end-to-end model with multiple conditioning factors. Instead of relying on a single end-to-end model, we explore to decompose the synthesis process into sequential stages, with each step focusing on a specific sub-task. This approach offers two advantages: \textbf{1)} it enables better dataset utilization by allowing different datasets to specialize in sub-tasks, \textbf{2)} improves factor separation by explicitly handling each attribute at distinct stages.

Building on this framework, we designed a three-stage pipeline for generating human images with arbitrary facial identity, clothing, pose and view as input (See Fig. \ref{fig:teaser}). The first stage generates a clothed A-pose human with a specified face and shoes, trained on the VTON dataset. Next, the second stage synthesizes the back view of the human using the MVHumanNet dataset. Finally, the third stage enables the generation of a human image in an arbitrary pose and viewpoint by combining the VTON and MVHumanNet datasets.
To condition human synthesis on pose and viewpoint control, we use the 3D parametric model SMPLX \cite{Pavlakos_2019_CVPR}, and render the SMPLX mesh from a given viewpoint and apply a UV-based texture encoding, where each body part is assigned a unique color-coded coordinate. With these structured controls, we implement a diffusion transformer (DiT)-based \cite{Peebles_2023_ICCV} architecture for back-view and free-view human synthesis , which integrates these disentangled factors into the synthesis pipeline. By incorporating these structured controls, our approach enables fine-grained manipulation of multiple human attributes within a unified framework. Our experiments demonstrate that, even when using a network structure similar to the end-to-end model, the stage-by-stage approach exhibits significantly better generalizability. This suggests that a stage-by-stage framework can more effectively utilize available data by assigning different datasets to specialized sub-tasks. We also observe that as the conditioning in the third stage becomes simpler, training by combining the MVHumanNet and VTON datasets leads to improved performance compared to training solely on the MVHumanNet dataset.

\begin{figure*}[htbp]
    \centering
    \includegraphics[width=\linewidth]{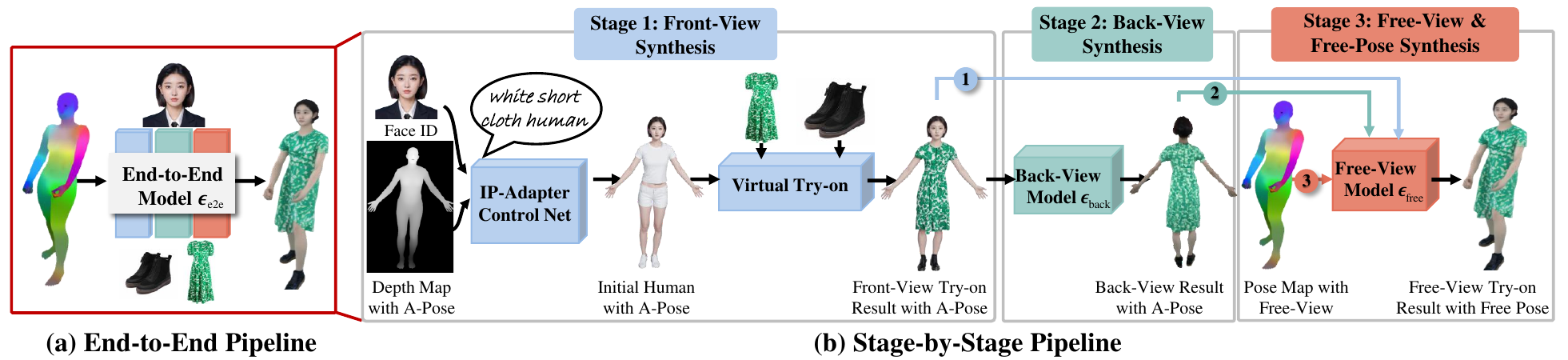}
    \vspace{-5mm}
    \caption{Overview of the proposed pipelines. (a) The end-to-end pipeline directly synthesizes the final image from disentangled inputs, including a face image, clothing images, and a pose map. (b) The stage-by-stage pipeline decomposes the process into three steps: front-view synthesis with identity and clothing control, back-view synthesis, and free-view synthesis under the target pose and viewpoint. Both pipelines are implemented using DiscoHuman, with details provided in Fig. \ref{fig:network}.}
    \label{fig:overview}
\end{figure*}

The contribution can be summarized as follows:
\begin{itemize}
    \item We introduce a new and challenging task in human image synthesis that aims to achieve explicit disentanglement and control over viewpoint, pose, clothing, and identity within a unified framework.
    \item We explore a novel end-to-end model that can address the disentangled and controllable human synthesis task and further propose a stage-by-stage framework that enhances control over pose and viewpoint, significantly improving generalizability, particularly for in-the-wild scenarios.
    \item Our experiments demonstrate that our methods achieve effective disentanglement, outperforming existing methods in view and pose control. We hope that our experiments and discussion on end-to-end versus stage-by-stage methods will inspire relevant human synthesis research.
\end{itemize}

\section{Related Work}
\label{sec:related}

\subsection{Human Synthesis}
In the realm of human synthesis, recent works \cite{dong2023ag3d,zhang2023getavatar,zhang2023styleavatar3d,fu2022stylegan,Fruhstuck_2022_CVPR,Sarkar2021HumanGAN,hong2023evad,xiong2023get3dhuman} have advanced the field by focusing on generating high-quality, full-body human images with diverse identities, hairstyles, garments, and poses. StyleGan-Human \cite{fu2022stylegan} adopts a data-centric perspective to examine the data factors influencing image generation based on the StyleGAN framework. InsetGAN \cite{Fruhstuck_2022_CVPR} integrates multiple pre-trained GANs, with each focusing on distinct components. To achieve control over pose, local body part appearance and garment style, HumanGAN \cite{Sarkar2021HumanGAN} is capable of performing all tasks related to global appearance sampling, pose transfer, part and garment transfer, as well as part sampling. EVA3D \cite{hong2023evad} and Get3DHuman \cite{xiong2023get3dhuman} expand the synthesis capabilities beyond 2D, pushing toward 3D human generation with realistic texture and pose control. EVA3D \cite{hong2023evad} incorporates a compositional human NeRF (Neural Radiance Field) representation, which divides the human body into 16 distinct parts and assigns each part an individual network. Each network models the corresponding local volume, allowing EVA3D to capture detailed textures and pose variations with high accuracy. Get3DHuman \cite{xiong2023get3dhuman}, lifting StyleGan-Human into 3D, advances full-body human synthesis and enables detailed control over body shape, garment, and even facial expression, making it highly applicable in applications like gaming and digital fashion.

\subsection{Pose Transfer}
With the rise of diffusion models, increasing research \cite{lu2024coarse,bhunia2023person,xu2024magicanimate,hu2024animate,shen2023advancing} has focused on realistic person image generation. PIDM \cite{bhunia2023person} first explores denoising diffusion for person synthesis, demonstrating high-quality results via iterative refinement.
PCDM \cite{shen2023advancing} adopts a step-by-step conditional framework to enhance pose control. MagicAnimate \cite{xu2024magicanimate} and AnimateAnyone \cite{hu2024animate} extend diffusion to person animation, achieving smooth pose transitions and dynamic visual effects.
Champ \cite{zhu2024champ} incorporates a 3D human model into latent diffusion for enhanced 3D conditioning. Human4DiT \cite{shao2024human4dit} leverages diffusion transformers to synthesize 360° coherent human videos from a single image, enabling fine-grained pose and view control.

\subsection{Virtual Try-on}
Image-based virtual try-on (VTON) synthesizes realistic images by transferring garment appearances onto their bodies. Recent methods use GANs or diffusion models to improve realism and precision. GAN-based approaches like VITON-HD \cite{Choi_2021_CVPR} address high-resolution image misalignment with ALIAS normalization, while HR-VITON \cite{lee2022high} integrates warping and segmentation into a unified try-on condition generator to reduce artifacts. However, GAN-based models often produce unnatural deformations in wild scenarios, reducing the fidelity and realism of the output result. In contrast, diffusion models  have recently demonstrated strong performance in image generation. OOTDiffusion \cite{xu2024ootdiffusion} aligns garment features with bodies using Outfitting Fusion, improving realism and controllability without redundant warping. IDM-VTON \cite{choi2025IDM} employs a visual encoder and parallel UNet for better garment encoding, while DressCode \cite{Morelli_2022_CVPR} introduces a multi-category dataset and a Pixel-level Semantic-Aware Discriminator (PSAD) to enhance image quality.

\section{Method}
\subsection{Overview}
Fig. \ref{fig:overview} provides an overview of the proposed pipelines for human image synthesis: an end-to-end pipeline and a stage-by-stage pipeline. Both pipelines leverage the proposed DiT-based model, \textbf{DiscoHuman}, which enables \textbf{dis}entangled and \textbf{co}ntrollable \textbf{human} image synthesis.  

The end-to-end pipeline directly generates a complete human image from a set of disentangled inputs, including a face image, clothing images (upper/lower clothing or a dress, and shoes), and a pose map, allowing simultaneous control over face identity, clothing, pose, and viewpoint.  In contrast, the stage-by-stage pipeline refines the synthesis process in three stages. It first generates an A-pose human image with identity and clothing control, then estimates the back view, and finally synthesizes a free-view image under a target pose and viewpoint. This decomposition enhances controllability and generalizability. DiscoHuman is adapted for the second and third stage by modifying its input configurations. Architectural details of DiscoHuman are further described in Sec. \ref{sec:disohuman}.  

\subsection{DiscoHuman}
\label{sec:disohuman}
\begin{figure*}[htbp]
    \centering
    \includegraphics[width=0.9\linewidth]{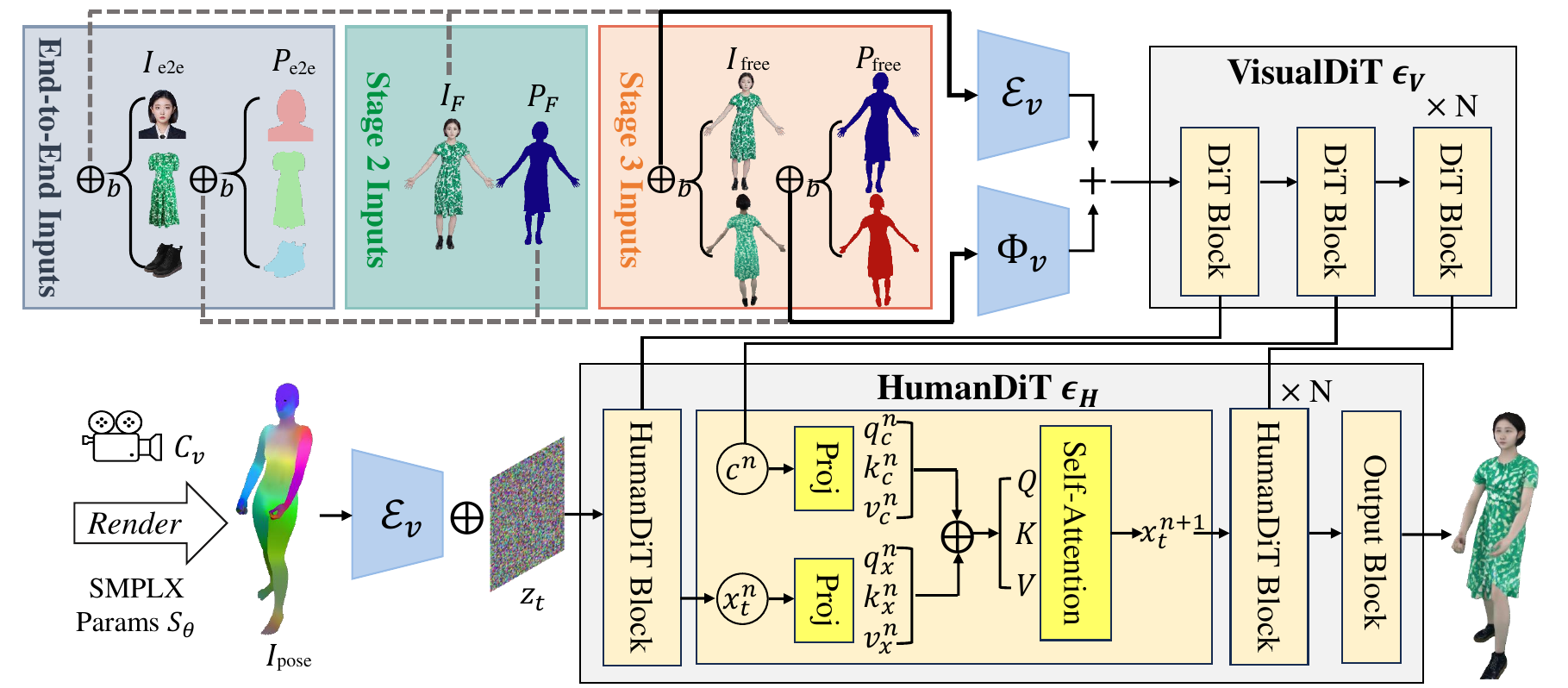}
    \caption{DiscoHuman model $ \epsilon $ consists of a VisualDiT $ \epsilon_V $ and a HumanDiT $ \epsilon_H $. The VisualDiT is responsible for encoding visual conditions, with different input settings depending on the pipeline or stage in which DiscoHuman is applied. The upper left blocks illustrate three possible input configurations. In this figure, the active configuration corresponds to Stage 3, while the inactive settings are indicated by grey dashed lines. To maintain simplicity, the denoising timestep $t$ is not shown in this figure.}
    \label{fig:network}
    \vspace{-2mm}
\end{figure*}

We developed DiscoHuman for disentangled and controllable human synthesis. Fig. \ref{fig:network} illustrates its structure. Given a set of disentangled conditions, it enables fine-grained control over appearance, pose, and viewpoint in human image generation. The model consists of two key components: VisualDiT Encoder, which encodes appearance-related conditions, and HumanDiT, which synthesizes the final human image while ensuring spatial alignment with the input pose.

\paragraph{VisualDiT Encoder}
The VisualDiT Encoder is a DiT-based module specifically designed for encoding visual conditions. Unlike standard DiT models, which take a noised latent as input, VisualDiT processes un-noised latents extracted from input images and conditions the denoising timestep at a fixed value $ t_0 $.

A key challenge in implementing VisualDiT lies in the variability of both the number and semantics of input conditions. For example, a sample may contain two garment images (e.g., top and bottom) or a single image (e.g., a dress). Designing separate encoders for each semantic category is computationally inefficient.
To overcome this, the VisualDiT encoder employs a unified strategy. Given a set of input images $ I = \{ I_1, I_2, ..., I_n \} $, each is encoded via a VAE encoder $ \mathcal{E}_v(\cdot) $ to produce latent representations. The corresponding segmentation masks $ P = \{ P_1, P_2, ..., P_n \} $ are one-hot encoded to indicate semantic categories. A convolutional embedding function $ \Phi_v(\cdot) $ processes these masks to align them with the image latents.

The VisualDiT Encoder then processes the concatenated image latents and embedded semantic maps, producing multi-level condition features $ c^n $ at each DiT block:
\begin{equation}
    c^n = \epsilon_V\left(\mathcal{E}_v\left(\oplus_b I\right) + \Phi_v\left(\oplus_b P\right), t_0\right),
\end{equation}
where the superscript $n$ indicates the index of the DiT block, $ \epsilon_V $ denotes VisualDiT, $ \mathcal{E}_v(\cdot) $ is the VAE encoder, and $ \oplus_b $ represents batch-wise concatenation.

\paragraph{HumanDiT}
\label{sec:humandit}

The HumanDiT model synthesizes the final human image based on the multi-level condition features $ c^n $ and a viewpoint-conditioned pose representation. To achieve precise pose and viewpoint control, we utilize the 3D parametric model SMPLX \cite{Pavlakos_2019_CVPR}. The SMPLX mesh is rendered under the target viewpoint, providing a structured pose representation that encodes both articulation and spatial alignment. The resulting pose map $ I_{\text{pose}} $ is then processed through the VAE encoder $ \mathcal{E}_v(\cdot) $ to ensure consistency with the visual condition features.

The HumanDiT model conditions the synthesis process on these features, formulated as:
\begin{equation}
\hat{\epsilon}_t = \epsilon_H \left(z_t \oplus \mathcal{E}_v(I_{\text{pose}}), t, c^n\right),
\end{equation}
where $ \epsilon_H $ represents the HumanDiT model, $ z_t $ is a random noise latent, and $ \oplus $ denotes channel-wise concatenation. The pose map $ I_{\text{pose}} $ is processed using the VAE encoder $ \mathcal{E}_v(\cdot) $, aligning its representation with the image features.

Within each DiT block of $ \epsilon_H $, effective feature fusion is essential. To achieve this, the diffusion feature $ x^n_t $ and the multi-level condition features $ c^n $ are separately projected into query ($ q $), key ($ k $), and value ($ v $) representations, which are concatenated in a length-wise manner:
\begin{equation}
\mathbf{Q} = q^n_{x} \oplus q^n_{c}, \quad
\mathbf{K} = k^n_{x} \oplus k^n_{c}, \quad
\mathbf{V} = v^n_{x} \oplus v^n_{c}.
\end{equation}
\begin{figure*}[htbp]
    \centering
    \includegraphics[width=0.98\linewidth, trim=8px 0 8px 0, clip]{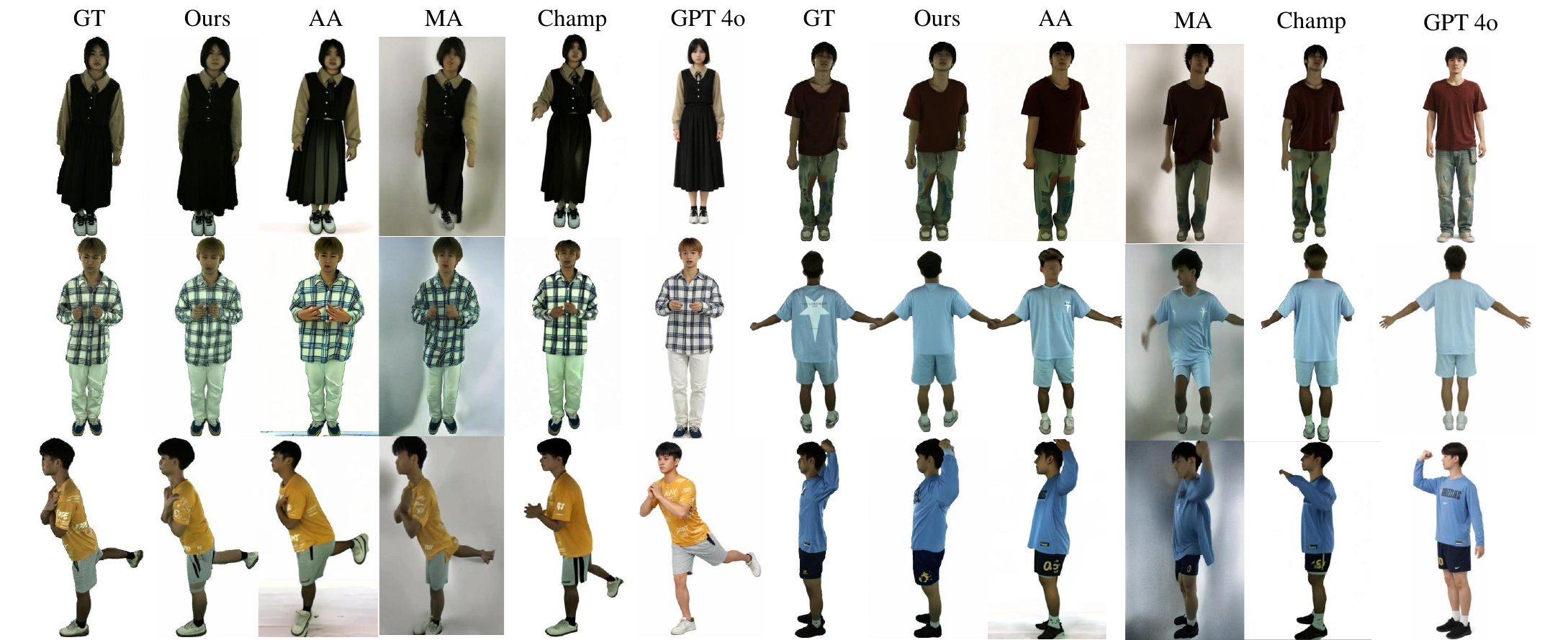}
    \caption{Qualitative comparison of different methods on MVHumanNet~\cite{Xiong_2024_CVPR}. We compare with AnimateAnyone (AA)~\cite{hu2024animate}, MagicAnimate (MA)~\cite{xu2024magicanimate}, Champ~\cite{zhu2024champ}, and GPT4o~\cite{hurst2024gpt}}
    \label{fig:MVHumanNet_animation_result}
\end{figure*}
\begin{figure*}[ht]
    \centering
    \includegraphics[width=0.92\linewidth]{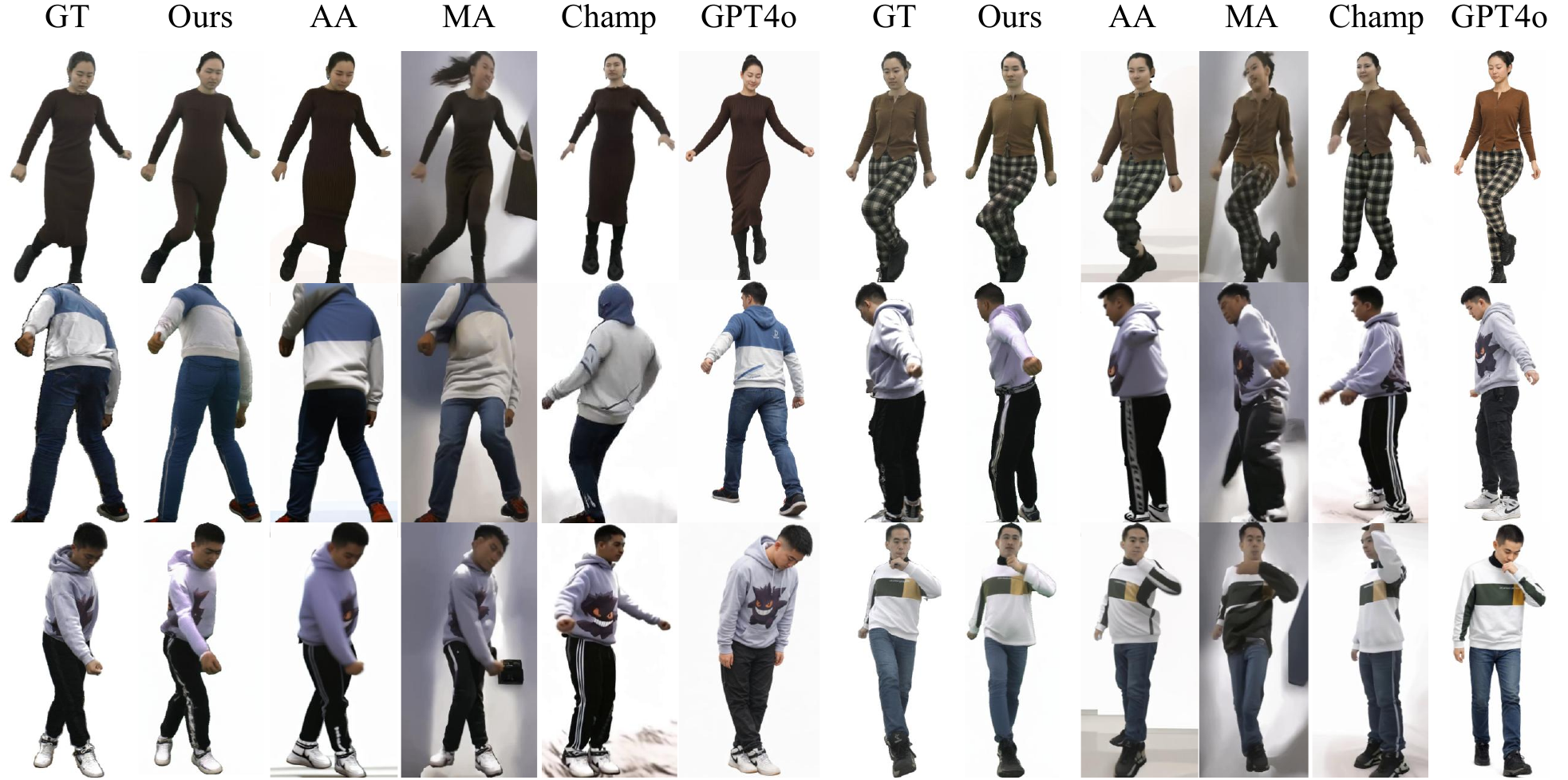}
    \caption{Qualitative comparison of different methods under the zero-shot setting on sampled results from AvatarReX~\cite{zheng2023avatarrex} (top) and THuman4.0~\cite{zheng2022structured} (bottom). We compare with AnimateAnyone (AA)~\cite{hu2024animate}, MagicAnimate (MA)~\cite{xu2024magicanimate}, Champ~\cite{zhu2024champ}, and GPT4o~\cite{hurst2024gpt}.}
    \label{fig:THuman_AvatarRex_animation_result}
\end{figure*}

A self-attention mechanism is then applied to fuse these features, allowing the HumanDiT model to effectively reference the conditioning information:
\begin{equation}
\hat{x}^{n+1}_t = \text{softmax}\left(\frac{\mathbf{QK}^T}{\sqrt{d}}\right) \mathbf{V}.
\end{equation}
Since concatenating $ c^n $ increases the sequence length, the excess portion is discarded after attention to maintain computational efficiency.

\paragraph{Rectified Flow Loss} To optimize the entire DiscoHuman model $\epsilon$, we applied rectified flow \cite{liu2022flow,esser2024scaling}. The forward and corresponding loss function are defined as follows:
\begin{subequations}
\begin{align}
    \hat{\epsilon}_t &= \epsilon \left(z_t, I_{\text{pose}}, t, I, P\right), \\
    \mathcal{L} &= \mathbb{E}_{t, x_0, x_t} \left[ w(t) \left\| \hat{\epsilon}_t - (x_t - x_0) \right\|^2 \right],
\end{align}
\end{subequations}
where $w(t)$ is a weighting function \cite{esser2024scaling}, $x_0$ is the latent representation of the ground truth image, and $x_t$ represents noised $x_0$ at timestep $t$.

\subsection{End-to-End (E2E) Pipeline}
In the end-to-end pipeline, we extract controllable factors from an A-pose image using SMPLX parameters and segmented face and clothing images. The inputs image and corresponding parsing maps to the VisualDiT Encoder are defined as:
\begin{subequations}
\begin{align}
    I_{\text{e2e}} &= \{ I_{\text{face}}, I_{\text{upper cloth}}, I_{\text{lower cloth}}, I_{\text{dress}}, I_{\text{shoes}} \}, \\
    P_{\text{e2e}} &= \{ P_{\text{face}}, P_{\text{upper cloth}}, P_{\text{lower cloth}}, P_{\text{dress}}, P_{\text{shoes}} \}.
\end{align}
\end{subequations}

The input settings for HumanDiT remain consistent with those described in Sec. \ref{sec:disohuman}. Since a person’s outfit consists of either separate garments (upper and lower) or a whole-body dress, we ensure consistency by inputting a black image with a semantic map labeled as pure `background' whenever a clothing category is missing. To train the e2e model, the forward process is defined as follows:
\begin{equation}
    \hat{\epsilon}_t = \epsilon_{\text{e2e}} \left(z_t\oplus \mathcal{E}_v(I_{\text{pose}}), t, I_{\text{e2e}}, P_{\text{e2e}}\right),
\end{equation}
where $\epsilon_{\text{e2e}}$ is the DiscoHuman model applied in end-to-end pipeline.
\subsection{Stage-by-Stage (SBS) Pipeline}
\paragraph{Stage 1: Front-View Synthesis} The first stage provides initial control over face identity and clothing. This process consists of two steps. First, IP-Adapter \cite{ye2023ip} is used to control face identity, while ControlNet \cite{Zhang_2023_ICCV} ensures that the human pose is fixed in an A-pose. A simple text prompt is used to generate a person wearing a tight white T-shirt and trousers, which simplifies the subsequent try-on process. Next, given the target clothing, IDM-VTON \cite{choi2025IDM} is applied to generate the try-on results. Since no dedicated model exists for shoe try-on, IP-Adapter with image-guided inpainting is used to apply shoes to the synthesized person.

\begin{table*}[htbp]
  \centering
   \caption{Quantitative evaluation on MVHumanNet \cite{Xiong_2024_CVPR}, THuman 4.0 \cite{zheng2022structured}, and AvatarReX dataset \cite{zheng2023avatarrex}. The comparison on THuman 4.0 and AvatarReX is conducted in a zero-shot manner. Cells highlighted in {\begin{tikzpicture} [scale=1.0] \fill[deeppeach] (0,0) rectangle (1em,0.8em); \end{tikzpicture}} {\begin{tikzpicture}[scale=1.0]  \fill[color=lightgray] (0ex,0ex) rectangle (1em,0.8em);  \end{tikzpicture} denotes the best and second-best performances.}}
  \label{tab:qualitative_comparison}

  \resizebox{\linewidth}{!}{
    \begin{tabular}{lcccccccccccc}
    \toprule
    \multicolumn{1}{c}{Dataset} & \multicolumn{4}{c}{MVHumanNet \cite{Xiong_2024_CVPR}} & \multicolumn{4}{c}{THuman 4.0 \cite{zheng2022structured}} & \multicolumn{4}{c}{AvatarReX \cite{zheng2023avatarrex}} \\
    \cmidrule(r{2pt}){1-1} \cmidrule(r{2pt}){2-5}
    \cmidrule(r{2pt}){6-9}
    \cmidrule(r{2pt}){10-13}
    \multicolumn{1}{c}{Method} & FID ↓ & LPIPS ↓ & PSNR ↑ & SSIM ↑ & FID ↓ & LPIPS ↓ & PSNR ↑ & SSIM ↑ & FID ↓ & LPIPS ↓ & PSNR ↑ & SSIM ↑ \\
    \cmidrule(r{2pt}){1-1} \cmidrule(r{2pt}){2-5}
    \cmidrule(r{2pt}){6-9}
    \cmidrule(r{2pt}){10-13}
    AnimateAnyone \cite{hu2024animate} & 70.7821  & 0.4331  & \cellcolor{lightgray} 13.2582  & 0.6750  & 74.1451 & 0.3642  & \cellcolor{lightgray} 14.7458  & \cellcolor{lightgray} 0.7799  & 69.2732  & 0.2157  & 15.5053  & 0.8264  \\
    MagicAnimate \cite{xu2024magicanimate} & 73.4553  & 0.4798  & 9.3189  & 0.7180  & 96.6649  & 0.4844  & 8.3149  & 0.6974  & 102.0733  & 0.5225  & 7.4551  & 0.6911  \\
    Champ \cite{zhu2024champ} & 62.8560 & \cellcolor{lightgray} 0.3994 & 13.1042 & \cellcolor{deeppeach} 0.8635 & 93.3094 & 0.4457 & 9.5117 & 0.7015 & 78.9367 & 0.3717 & 10.4287 & 0.7676 \\
    \cdashline{1-13}
    Ours (E2E) & \cellcolor{lightgray} 25.1648 & 0.4494 & 10.7573 &  0.6606 & \cellcolor{lightgray} 69.3216 & \cellcolor{lightgray} 0.3582 & 12.7761 & 0.7646 & \cellcolor{lightgray} 63.0782 & \cellcolor{lightgray} 0.2091 & \cellcolor{lightgray} 15.6723 & \cellcolor{lightgray} 0.8312 \\
    Ours (SBS)  & \cellcolor{deeppeach} 14.4210  & \cellcolor{deeppeach} 0.1622  & \cellcolor{deeppeach} 17.0370  & \cellcolor{lightgray} 0.8311  & \cellcolor{deeppeach} 58.5134  & \cellcolor{deeppeach} 0.2262  & \cellcolor{deeppeach} 14.7664  & \cellcolor{deeppeach} 0.7870  & \cellcolor{deeppeach} 53.0746  & \cellcolor{deeppeach} 0.1744  & \cellcolor{deeppeach} 16.1607  & \cellcolor{deeppeach} 0.8378  \\
    \bottomrule
    \end{tabular}
    }
\end{table*}

\begin{figure*}[htbp]
    \centering
    \includegraphics[width=0.98\linewidth]{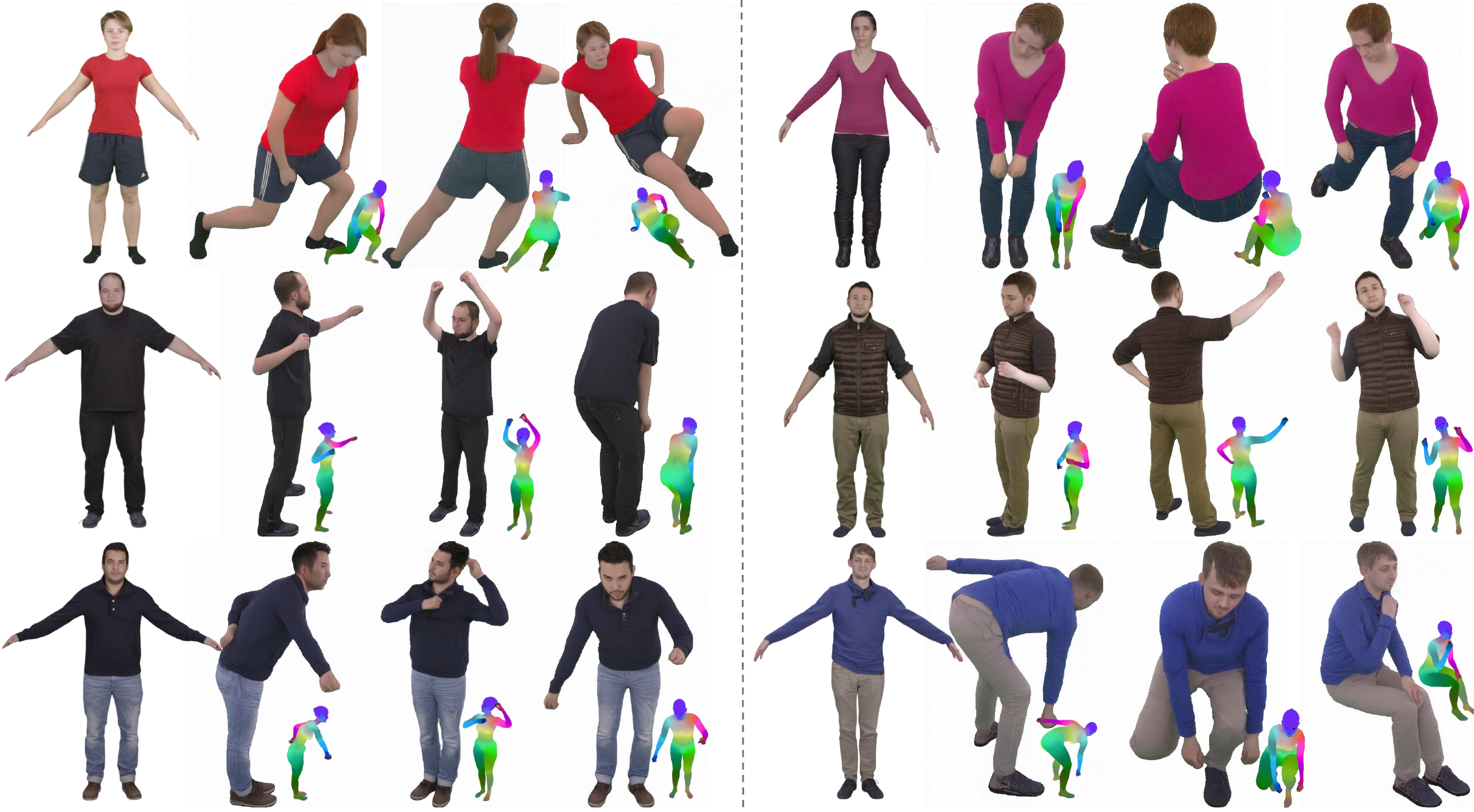}
    \caption{Given a reference person image from the People Snapshot dataset \cite{alldieck2018video} (leftmost in each sample) and a target pose (visualized as a colored SMPLX body map), our method (Stage-by-Stage pipeline) synthesizes the same person performing the new action from a specified novel view. Note that our model is evaluated in a zero-shot setting on this dataset, demonstrating strong generalization in pose transfer and view synthesis.}
    \label{fig:people_snapshop}
\end{figure*}

\paragraph{Stage 2: Back-View Synthesis}
The second stage synthesizes a realistic back-view image $I_B$ of the previous generated front-view image $I_F$. This is achieved by reusing the DiscoHuman model with modified inputs in the VisualDiT Encoder, where the front-view image serves as the visual condition. Since only the front image is available, the corresponding semantic map is derived from its foreground segmentation as $P_F$. The back-view generation model is formulated as follows:
\begin{equation}
\hat{\epsilon}_t = \epsilon_\text{back} \left(z_t, I_{\text{A-pose}}, t, I_F, P_F\right).
\end{equation}
During training, we observe significant lighting shifts between the front and back camera views in the MVHumanNet dataset. Directly using raw MVHumanNet images introduces color saturation shifts, leading to unrealistic textures in the synthesized back views. 
To mitigate this issue, we employ the FLUX \cite{flux2024} model to enhance the training image quality, particularly in frontal and back A-pose images. As demonstrated in our ablation study in Sec. \ref{sec:ablation}, this enhancement significantly improves synthesis quality.

\paragraph{Stage 3: Free-View \& Pose Synthesis} The final stage generates a human image under an arbitrary pose and viewpoint using the outputs from Stage 1 and Stage 2 along with a pose map. To obtain the foreground segmentation map of $I_F$ and $I_B$, SAM model \cite{Kirillov_2023_ICCV} is applied to obtain $P_F$ and $P_B$. Thus we have $I_\text{free}=\{I_F, I_B\}$, $P_\text{free}=\{P_F, P_B\}$. The free-view synthesis model is formulated as follows:
\begin{equation}
\hat{\epsilon}_t = \epsilon_\text{free} \left(z_t, I_{\text{pose}}, t, I_\text{free}, P_\text{free}\right).
\end{equation}

\section{Experiments}
\subsection{Implementation Details}
In the implementation of DiscoHuman, both VisualDiT and HumanDiT utilize the pre-trained DiT model from Stable Diffusion 3 Medium \cite{esser2024scaling}, which consists of 24 DiT blocks. To conserve GPU memory, VisualDiT employs only the first 12 blocks. The corresponding block features $c^n$ from VisualDiT are duplicated across all blocks to ensure alignment with HumanDiT. The synthesis resolution for the image is set to $512\times 768$. For hyperparameter settings, we use an 8-bit Adam optimizer with a learning rate of $2 \times 10^{-5}$ and a batch size of 16. All experiments are conducted on two A100 GPUs for 30k training steps. 

\begin{figure*}[htbp]
    \centering
    \includegraphics[width=0.98\linewidth]{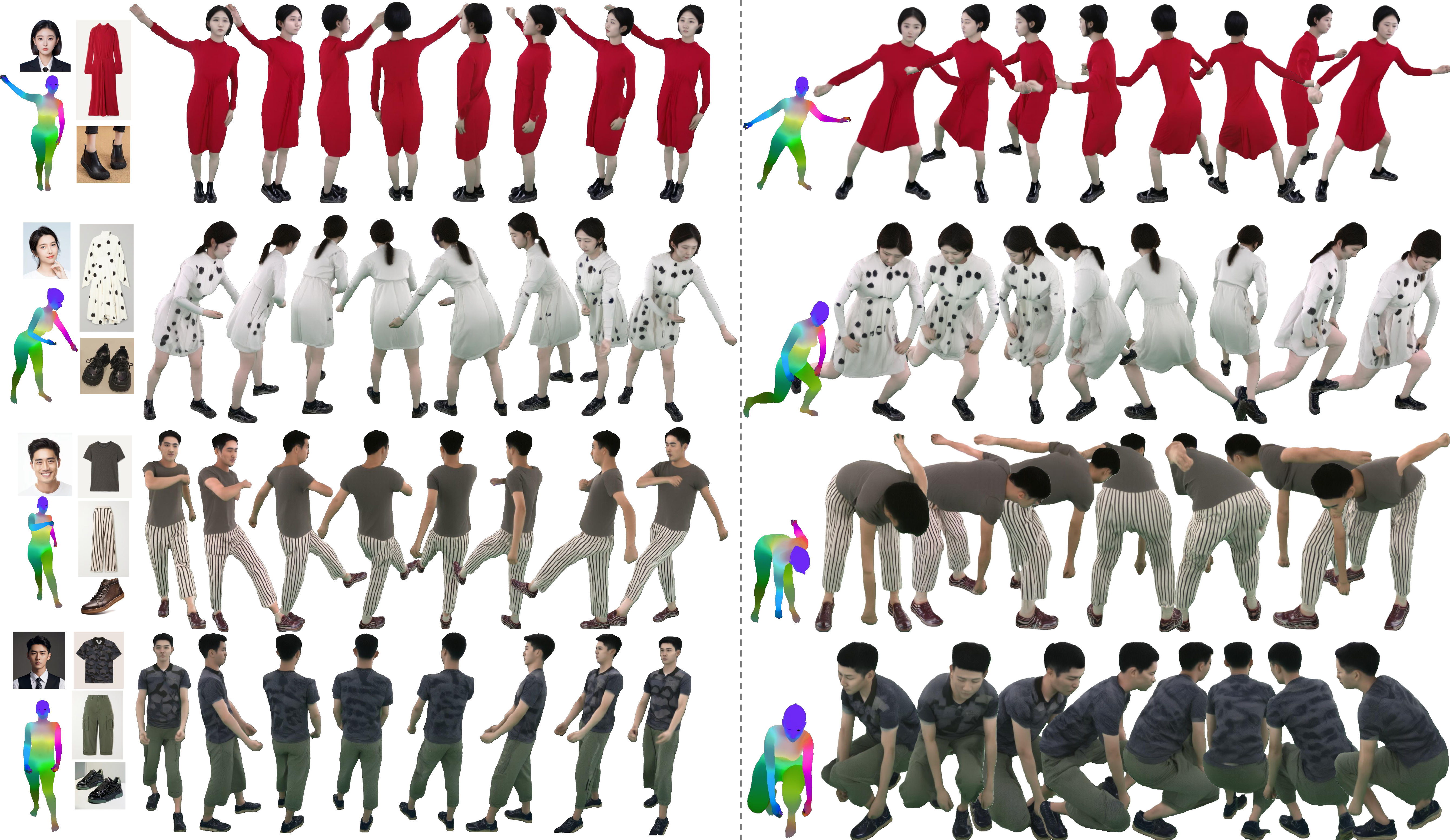}
    \caption{View interpolation results achieved by the Stage-by-Stage (SBS) pipeline. Each row shows synthesized images of the same identity and clothing under two poses and multiple viewpoints. Despite being an image generation model, DiscoHuman produces smooth and consistent transitions across views, demonstrating strong continuity and factor disentanglement.}
    \label{fig:pose_view-interpolation}
\end{figure*}

\begin{figure}[t]
    \centering
    \includegraphics[width=0.98\linewidth]{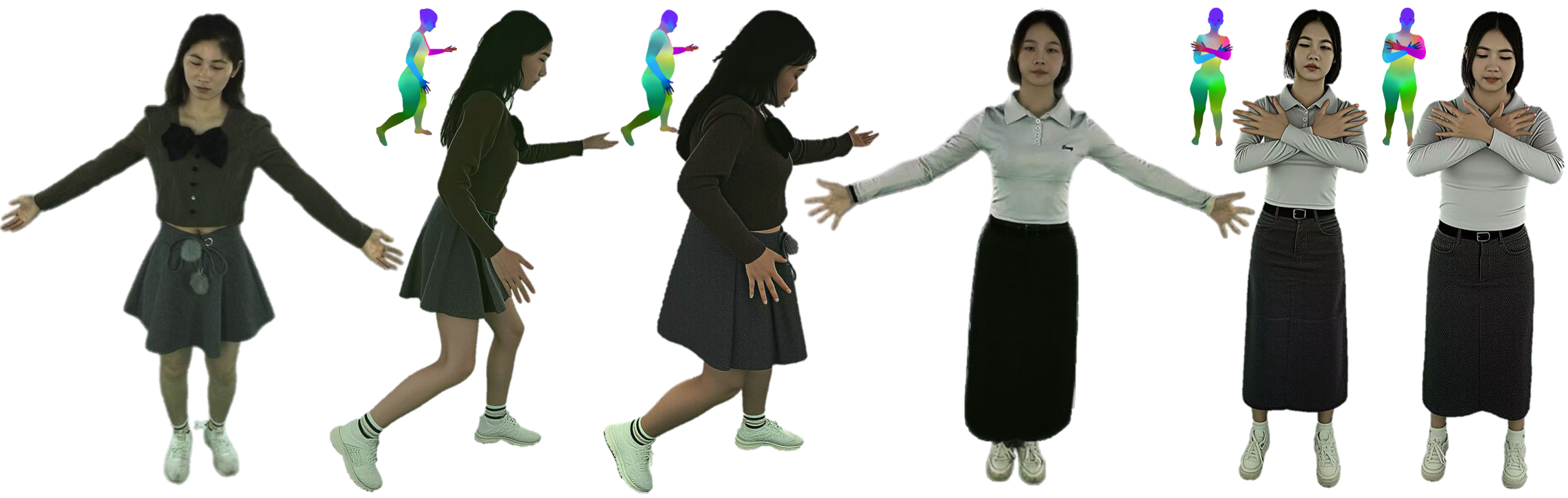}
    \caption{Shape control under varying shape parameters. The leftmost image in each example is the input, and the corresponding pose is shown in the top-left corner of each result.}
    \label{fig:shape_analysis}
\end{figure}

\subsection{Datasets Preparation}
\label{sec:dataset}
\paragraph{MVHumanNet} The MVHumanNet dataset \cite{Xiong_2024_CVPR} consists of multi-view videos of 9,000 subjects wearing various outfits. It provides annotations for human foreground masks, camera poses, and SMPLX estimations. Each subject includes a calibrated A-pose frame and is captured by multi-view cameras. To ensure data quality, we use only the 6,000 high-quality subjects, as some subjects have inconsistent camera settings. Images are extracted from video frames, and to reduce the dataset size, we select 8 evenly spaced frames per video and 4 specific camera views (front, left, back, right) for training, yielding a total of 192k training images. For testing, the dataset provides 400 test subjects, from which we select 2 frames per video and 4 camera views per subject, resulting in 3.2k test images.

For training the end-to-end pipeline, we extract face, upper clothing, lower clothing, dress, and shoes from each subject's calibrated A-pose frame using the Sapiens segmentation model \cite{khirodkar2025sapiens}. The A-pose frame is not included in the training set but is used to provide disentangled inputs. To support more customizable inputs, the end-to-end pipeline is trained with images from both front and back views. For simplicity, these additional views are not depicted in Fig. \ref{fig:overview} and Fig. \ref{fig:network}.

\paragraph{VTON} The term "VTON dataset" in this paper refers to a combination of two virtual try-on datasets: VITON-HD \cite{Choi_2021_CVPR} and DressCode \cite{Morelli_2022_CVPR}. Both datasets provide human parsing annotations and flattened in-shop clothing images. We use PyMAF-X \cite{zhang2023pymaf} for SMPLX estimations. The training set consists of 60k images. Since these datasets contain only limited viewpoints and poses, we do not use them for model evaluation.

\paragraph{THuman 4.0} THuman 4.0 \cite{zheng2022structured} contains only three subjects. Due to its limited size, we use this dataset exclusively for zero-shot testing.

\paragraph{AvatarReX} AvatarReX \cite{zheng2023avatarrex} provides multi-view videos of four subjects. We select around 900 frames in total from THuman4.0 and AvatarRex to construct a zero-shot test set.

\begin{figure*}[htbp]
    \centering
    \begin{minipage}[c]{0.43\linewidth}
        \centering
        \includegraphics[height=230pt,width=\linewidth]{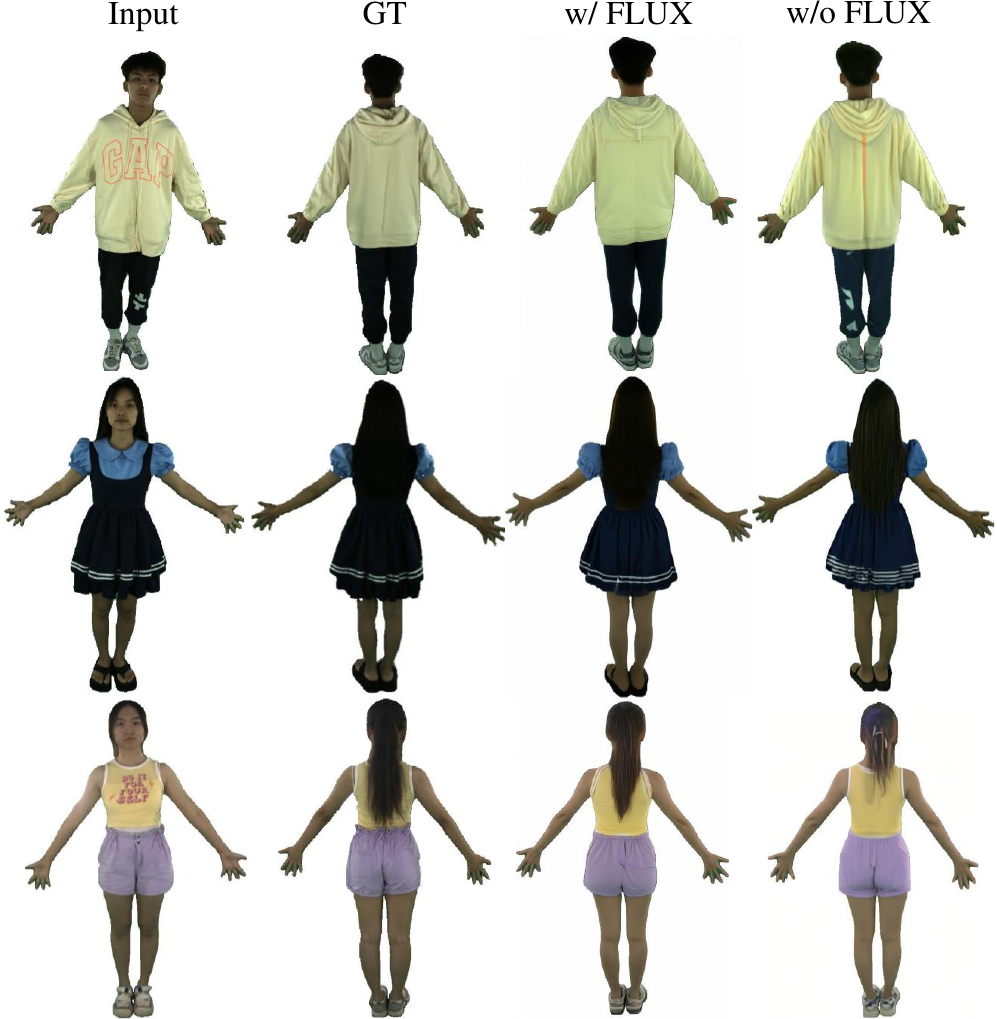}
        \caption{Ablation results of back-view synthesis models trained with and without FLUX-enhanced data.}
        \label{fig:back_view_ablation}
    \end{minipage}
    \hfill
    \begin{minipage}[c]{0.56\linewidth}
        \centering
        \includegraphics[height=230pt,width=\linewidth]{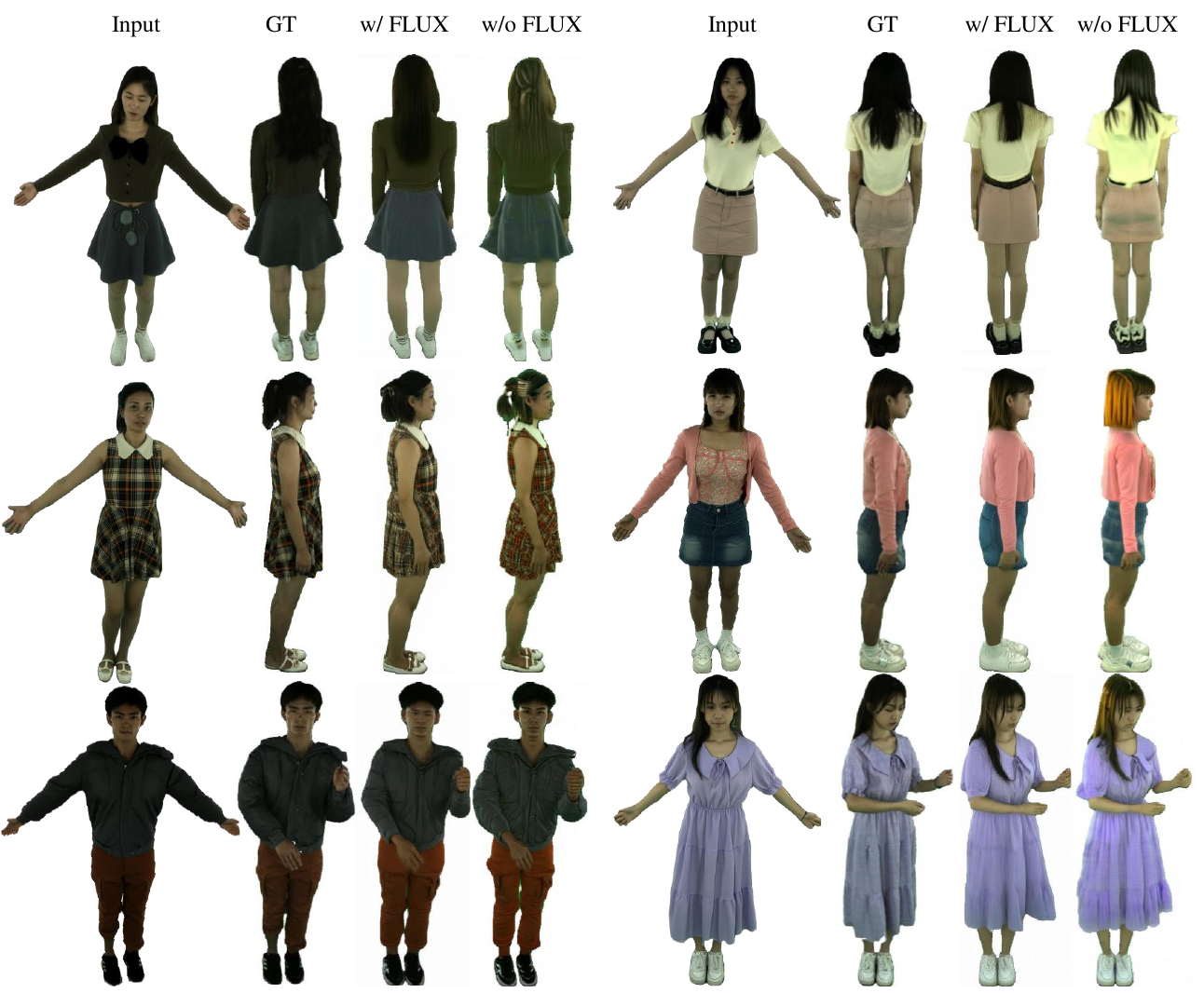}
        \caption{Free-view synthesis results based on back-view models trained with and without FLUX-enhanced data.}
        \label{fig:free_view_ablation}
    \end{minipage}
\end{figure*}

\begin{table*}[ht]
    \centering
    \begin{minipage}[c]{0.49\linewidth}
        \centering
    \caption{Ablation of FLUX enhancement for back-view training in the stage-by-stage pipeline. Best results are highlighted in \begin{tikzpicture}[scale=1.0] \fill[deeppeach] (0,0) rectangle (1em,0.8em); \end{tikzpicture}.}
    \label{tab:ablation}
    \resizebox{0.9\linewidth}{!}{
    \begin{tabular}{cccccc}
    \toprule
    Method & FLUX & FID ↓ & LPIPS ↓ & PSNR ↑ & SSIM ↑ \\
    \midrule
    Back-View Model & \checkmark   & \cellcolor{deeppeach} 38.5276  & 0.1852  & 15.3949  & 0.8077  \\
    Back-View Model &   & 46.0028  & \cellcolor{deeppeach} 0.1723  & \cellcolor{deeppeach} 15.6032  & \cellcolor{deeppeach}0.8179  \\
    \midrule
    Free-View Model & \checkmark  & \cellcolor{deeppeach} 20.8115  & \cellcolor{deeppeach} 0.1711  & \cellcolor{deeppeach} 16.9394  & \cellcolor{deeppeach} 0.8275  \\
    Free-View Model &   &  27.2246  & 0.1790  & 15.9477  & 0.8090  \\
    \bottomrule
    \end{tabular}
    }
    \end{minipage}
    \hfill
    \begin{minipage}[c]{0.49\linewidth}
    \centering
    \caption{Ablation study on feature representation and fusion strategies. The best results are highlighted in \begin{tikzpicture}[scale=1.0] \fill[deeppeach] (0,0) rectangle (1em,0.8em); \end{tikzpicture}.}
    \label{tab:ablation_modules}
  \resizebox{\linewidth}{!}{
  \begin{tabular}{lcc cccc}
    \toprule
    Variant & Level & Fusion & FID ↓ & LPIPS ↓ & PSNR ↑ & SSIM ↑ \\
    \cmidrule(lr){1-3} \cmidrule(lr){4-7}
    Direct injection & -- & Add 
    & 108.2222 & 0.3525 & 14.7477 & 0.7654 \\
    
    Single-level & Single & Self-attn 
    & 18.1163 & 0.1665 & 16.0658 & 0.8223 \\
    
    Multi + cross & Multi & Cross-attn 
    & 26.5646 & 0.1701 & 15.3487 & 0.8138 \\
    
    \textbf{Ours} & Multi & Self-attn 
    & \cellcolor{deeppeach} 14.4210 
    & \cellcolor{deeppeach} 0.1622 
    & \cellcolor{deeppeach} 17.0370 
    & \cellcolor{deeppeach} 0.8311 \\
    \bottomrule
  \end{tabular}
  }
    \end{minipage}
 \vspace{-5px}
\end{table*}

\subsection{Comparison}
Our goal is to address a human image synthesis task with disentangled control over multiple factors, including identity, clothing, pose, and viewpoint. To tackle this problem, we adopt a Stage-by-Stage (SBS) pipeline, which decomposes the generation process into multiple stages. As discussed in Sec.~\ref{sec:e2e_vs_sbs}, this design offers advantages over End-to-End (E2E) formulations in terms of disentanglement and controllability. Therefore, we focus our evaluation on the proposed SBS pipeline in this section.

To the best of our knowledge, no prior work fully addresses this problem setting. To enable meaningful comparison, we restrict our evaluation to the combined Stage 2 and Stage 3 pipeline, which corresponds to the widely studied pose-guided human image synthesis setting and allows for direct comparison with existing methods. Under this setting, we further adopt a unified input protocol in which all methods take the same canonical front-view image as input, ensuring fair and controlled comparisons.

We compare our method with representative state-of-the-art approaches, including AnimateAnyone \cite{hu2024animate}, MagicAnimate \cite{xu2024magicanimate}, and Champ \cite{zhu2024champ}. Since the official implementation of AnimateAnyone is unavailable, we use a third-party reimplementation\footnote{Moore-AnimateAnyone: \url{https://github.com/MooreThreads/Moore-AnimateAnyone}}. We additionally include GPT-4o \cite{hurst2024gpt} for qualitative comparison only.

\paragraph{Qualitative Evaluation}

Fig.~\ref{fig:MVHumanNet_animation_result} shows qualitative results on MVHumanNet. Our method produces consistent results under challenging pose and view changes, while baseline methods often fail to maintain plausible body structure or detailed appearance.

GPT-4o \cite{hurst2024gpt} is evaluated using its multi-image prompting capability, where we provide both front/back views and rendered SMPL-X pose maps as inputs. Despite leveraging these additional cues, GPT-4o primarily focuses on visual realism and often fails to maintain consistency in identity, pose, and viewpoint, leading to structural mismatches in challenging cases.

We further evaluate zero-shot generalization on THuman 4.0~\cite{zheng2022structured} and AvatarReX~\cite{zheng2023avatarrex} (Fig.~\ref{fig:THuman_AvatarRex_animation_result}), where all methods are tested without additional training. Our method maintains stable pose-guided synthesis and identity consistency across unseen data.

While Champ achieves competitive texture quality, it is less robust under complex pose and view changes. AnimateAnyone and MagicAnimate also show limitations in preserving fine-grained clothing details.

\paragraph{Quantitative Evaluation}

Tab.~\ref{tab:qualitative_comparison} reports the quantitative results using FID, LPIPS, PSNR, and SSIM. Our method achieves the best overall performance across datasets, outperforming prior approaches on most metrics.

Notably, results on THuman 4.0 and AvatarReX are obtained in a zero-shot setting, demonstrating strong generalization ability. While AnimateAnyone achieves comparable performance on PSNR and SSIM, it yields significantly worse FID, indicating inferior perceptual quality.

On MVHumanNet, Champ obtains slightly higher SSIM, reflecting better structural similarity. However, our method achieves superior performance in FID and LPIPS. While Champ obtains slightly higher SSIM, it tends to show weaker consistency across poses and viewpoints. In contrast, our method remains more stable under varying view settings, demonstrating stronger robustness in handling view and pose variations.

\begin{figure}[t]
    \centering
    \includegraphics[width=0.99\linewidth]{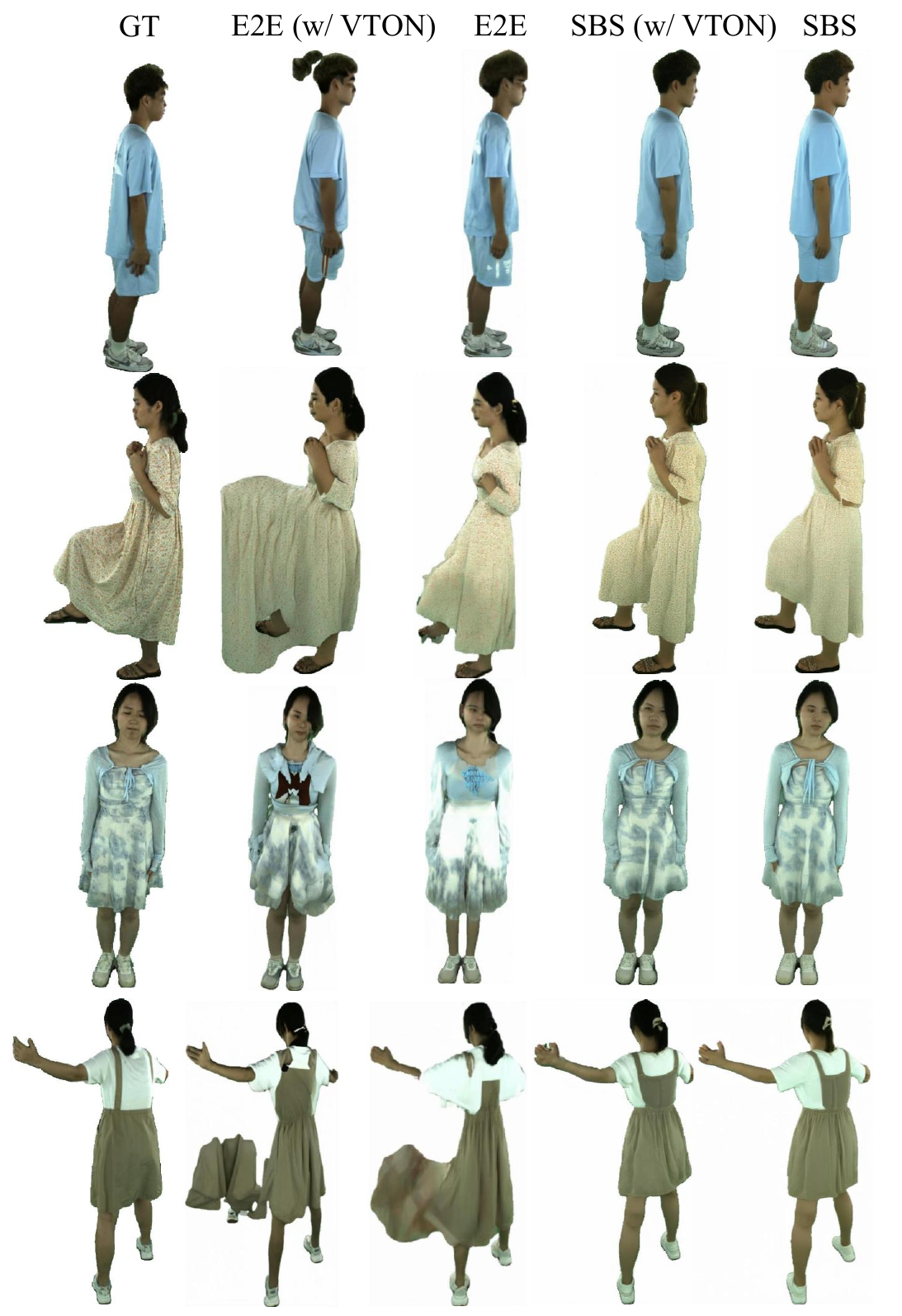}
    \caption{Qualitative comparison for End-to-End (E2E) and Stage-by-Stage (SBS) results on MVHumanNet~\cite{Xiong_2024_CVPR} dataset.}
    \label{fig:discussion}
\end{figure}

\subsection{Generalization and Controllability Analysis}

\paragraph{Zero-shot Generalization} Additional zero-shot results on People Snapshot~\cite{alldieck2018video} (Fig.~\ref{fig:people_snapshop}) further demonstrate the generalization capability of our approach. We attribute this robustness to the stage-by-stage design, which decouples appearance modeling and pose control, enabling more stable performance on unseen subjects and data distributions.

\paragraph{Full Pipeline Results} We also present full-pipeline results integrating all three stages of our framework (Fig.~\ref{fig:teaser} and Fig.~\ref{fig:pose_view-interpolation}), 
demonstrating its ability to achieve disentangled and controllable human image synthesis in more general and unconstrained settings, enabled by the proposed stage-by-stage design.

\paragraph{Body Shape Control} Fig.~\ref{fig:shape_analysis} shows that varying the SMPL-X shape parameters allows our method to reflect body shape to some extent. However, while our method allows for coarse control over body shape, achieving fine-grained shape control remains challenging due to the limited diversity of body shapes in the training data. More precise shape control is left for future work.

\subsection{Ablation Study}
\label{sec:ablation}
\paragraph{Effect of FLUX-based Data Enhancement}
During Stage 2 Back-View Synthesis, we find that directly using a front-view A-pose image to predict its back-view counterpart is highly challenging. This difficulty arises not only from the missing information on the back but also from the inconsistent lighting conditions in the MVHumanNet dataset. As a result, if we explicitly use the synthesized back-view results in Stage 3, we observe a notable degradation in performance compared to the training setup, which uses a ground-truth A-pose back-view image as input.

To address this, we conduct an ablation study showing that improving the quality of front-back A-pose images significantly benefits both back-view synthesis and free-view generation. As shown in Tab. \ref{tab:ablation}, applying FLUX \cite{flux2024} to enhance training A-pose images improves FID, with marginal changes in LPIPS, PSNR, and SSIM. Notably, models trained on FLUX-enhanced data ($\checkmark$ in the FLUX column) yield consistent gains across all metrics when generating free-view images from synthesized back views. As shown in Fig. \ref{fig:back_view_ablation} and Fig. \ref{fig:free_view_ablation}, training with FLUX-enhanced data significantly improves the quality of generated hair and clothing textures.
\paragraph{Architecture Design Ablation}
Tab.~\ref{tab:ablation_modules} reports quantitative ablation results for different design choices in our module, focusing on feature level and fusion strategy.

A direct injection baseline, conceptually similar to ControlNet-style designs, is first considered. In this setting, reference image features are injected into the diffusion process in a multi-level manner using a simple additive fusion. As shown in the first row, this strategy leads to a significant degradation across all metrics, indicating that direct feature injection, even with multi-level signals, is insufficient for capturing detailed human appearance.

The effect of feature hierarchy is further examined. The \textit{Single-level} variant only uses the feature from the final block of the encoder, while our method aggregates features across multiple levels. As shown in the second row, removing multi-level features results in noticeable performance drops, demonstrating that hierarchical representations provide complementary information for modeling fine-grained human textures.

Different feature fusion strategies are also evaluated. The \textit{Multi + cross} variant replaces our self-attention-based fusion with a cross-attention mechanism. As shown in the third row, this modification leads to inferior performance, suggesting that cross-attention is less effective for integrating appearance features with diffusion features in our setting.

Our full model combines multi-level feature representation with self-attention-based fusion, achieving the best performance across all metrics. These results highlight the importance of both hierarchical feature modeling and appropriate fusion design in controllable human image synthesis.

\section{User Study}
We conducted a user study to evaluate the effectiveness of our method on in-the-wild data. We randomly selected 30 cases, each containing four synthesized images generated by the four models under the same input conditions. A total of 50 participants were asked to vote for the image they perceived as the most realistic by considering: view consistency, pose accuracy, identity consistency, and texture fidelity. The results are shown in Tab. \ref{tab:user_study}.

\begin{table}[t]
    \centering
    \caption{User study results on in-the-wild data. The table shows the percentage of votes received by each method, where a higher percentage indicates stronger user preference.}
    \small
    \resizebox{0.85\linewidth}{!}{
    \begin{tabular}{cccc}
        \toprule
        Our Method & AnimateAnyone & MagicAnimate & Champ \\
        \midrule
        61.2\% & 10.8\% & 6.6\% & 21.4\%\\
        \bottomrule
    \end{tabular}
    }
    \label{tab:user_study}
\end{table}

\begin{table}[t]
  \centering
  \caption{Quantitative results of end-to-end (E2E) vs. stage-by-stage (SBS) training under different dataset combinations. We mark the best and second-best results with \begin{tikzpicture} [scale=1.0] \fill[deeppeach] (0,0) rectangle (1em,0.8em); \end{tikzpicture} \begin{tikzpicture}[scale=1.0]  \fill[color=lightgray] (0ex,0ex) rectangle (1em,0.8em);  \end{tikzpicture}. }
    \resizebox{0.95\linewidth}{!}{
    \begin{tabular}{llcccc}
    \toprule
    \multicolumn{1}{c}{Method} & \multicolumn{1}{c}{Training Dataset} & FID ↓ & LPIPS ↓ & PSNR ↑ & SSIM ↑ \\
    \midrule
    E2E   & MVHumanNet + VTON & 25.1648  & 0.4494  & 10.7573  & 0.6606  \\
    E2E   & MVHumanNet & \cellcolor{lightgray} 20.7037  & 0.3747  & 12.5632  & 0.7294  \\
    \cdashline{1-6}
    SBS & MVHumanNet + VTON & \cellcolor{deeppeach} 14.4210  & \cellcolor{deeppeach} 0.1622  & \cellcolor{deeppeach} 17.0370  & \cellcolor{deeppeach} 0.8311 \\
    SBS & MVHumanNet & 20.8115  & \cellcolor{lightgray} 0.1711  & \cellcolor{lightgray} 16.9394  & \cellcolor{lightgray}  0.8275 \\
    \bottomrule
    \end{tabular}
    }
  \label{tab:discussion}
\end{table}

\section{Discussion: End-to-End vs. Stage-by-Stage}
\label{sec:e2e_vs_sbs}

Our results suggest that directly training a unified end-to-end (E2E) model for controllable human synthesis is suboptimal when supervision comes from heterogeneous datasets. Instead, decomposing the task into a stage-by-stage (SBS) pipeline leads to more stable and effective learning.

We attribute the limitation of E2E to inconsistent conditioning across datasets. MVHumanNet provides rich control signals (e.g., identity, clothing, pose, and viewpoint), while VTON lacks several factors such as face and viewpoint. Filling missing inputs with $\varnothing$ using classifier-free guidance~\cite{ho2022classifier} introduces a distribution gap, which disrupts joint training and leads to degraded performance.

In contrast, SBS alleviates this issue by decoupling the generation process. By restricting cross-dataset training to Stage 3 with simplified conditioning (pose and texture), the model avoids entangling incompatible factors and reduces cross-dataset interference. This enables more effective utilization of heterogeneous data sources.

This advantage is reflected in both qualitative and quantitative results. As shown in Fig.~\ref{fig:discussion}, SBS produces more coherent textures and fewer artifacts, especially when incorporating VTON data. Tab.~\ref{tab:discussion} further shows that SBS improves LPIPS, PSNR, and SSIM compared to E2E, despite a slight drop in FID.

\section{Ethical Considerations}

Our work explores controllable human image synthesis, which can benefit applications such as virtual try-on and digital content creation. At the same time, such technologies may be misused to generate misleading or manipulated human images, especially when involving identity and appearance control. 

To reduce potential risks, our method relies on structured inputs rather than generating identities freely, which provides a degree of user control and traceability. We also note that the model is trained on existing datasets, which may introduce biases in body shape, clothing, and capture conditions, and thus may not generalize equally across all scenarios.

We emphasize that this work is intended for benign and creative applications, and encourage responsible use with appropriate safeguards such as user consent and content moderation.

\section{Limitations and Future Work}
One limitation of our approach lies in the diffusion model’s tendency to produce artifacts in smaller regions of the image, such as fingers and faces, due to their relatively small size compared to the entire picture. Consequently, the synthesized results from DiscoHuman occasionally exhibit imperfections in these areas. However, this limitation can be mitigated through post-processing methods that do not require additional fine-tuning. For example, utilizing the ADetailer plugin\footnote{ADetailer plugin: \url{https://github.com/Bing-su/adetailer}} effectively addresses these artifacts. This plugin re-generates the face and hand regions with a larger cropped resolution and seamlessly pastes the improved regions back into the original image, significantly enhancing the overall quality. Future work could explore integrating such post-processing techniques directly into the generation pipeline for a more streamlined and robust output.

\section{Conclusion}
We introduce a new and challenging task in human image synthesis that controls viewpoint, pose, clothing, and identity within a unified framework. While an end-to-end model offers a straightforward approach, our analysis reveals its limitations in generalizability, particularly when trained on datasets with inconsistent conditioning factors. To address this, we propose a stage-by-stage framework, which significantly enhances pose and viewpoint control while improving robustness to in-the-wild scenarios. Our experiments demonstrate that structured factor disentanglement leads to improved synthesis quality, particularly in zero-shot settings. By leveraging dataset specialization in different stages, our method enables more effective utilization of available data, outperforming existing approaches in view and pose control. However, our approach is still limited by the diffusion model’s tendency to produce artifacts in small regions (e.g., fingers and faces) due to their relative size. We will resolve this in future work.
We hope our findings on end-to-end vs. stage-by-stage synthesis will inspire further research in controllable human generation.

\bibliographystyle{IEEEtran}
\bibliography{IEEEabrv,main}
\vspace{-50pt}
\begin{IEEEbiography}[{\includegraphics[width=1in,height=1.25in,clip,keepaspectratio]{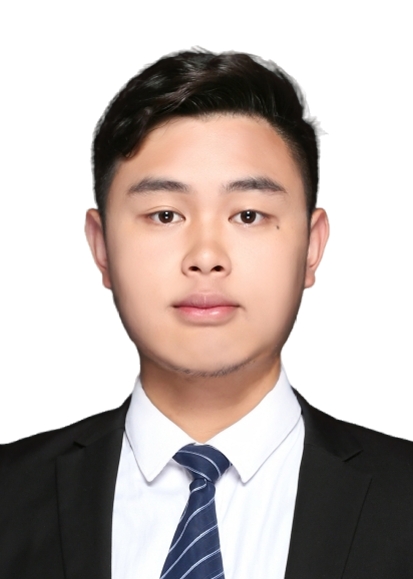}}]{Zhengwentai Sun}
 received his Bachelor's degree from the University of Electronic Science and Technology of China in 2021. He obtained a Master of Philosophy degree from The Hong Kong Polytechnic University in 2024. He is pursuing a PhD degree at The Chinese University of Hong Kong, Shenzhen. His research interests include image generation, editing, and their applications in human-centric contexts.
\end{IEEEbiography}
\vspace{-40pt}

\begin{IEEEbiography}[{\includegraphics[width=1in,height=1.25in,clip,keepaspectratio]{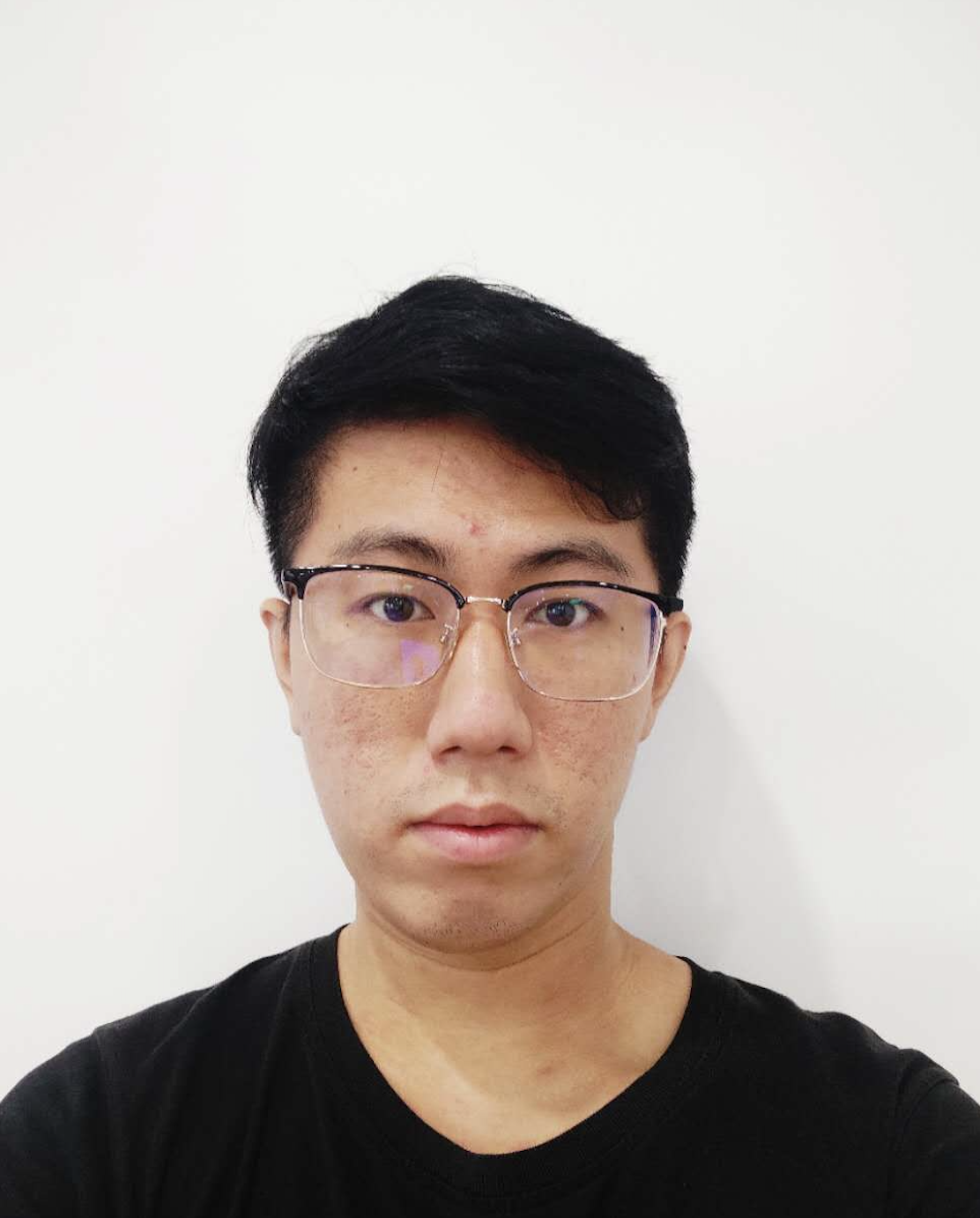}}]{Chenghong Li} is currently pursuing the PhD degree at The Chinese University of Hong Kong, Shenzhen. He received his the B.E. degree in 2018 from Jilin University and Master degree in 2021 from Xi'an Jiaotong University. His research interests include
computer vision and computer graphics, with a focus on human reconstruction and animation.
\end{IEEEbiography}
\vspace{-40pt}

\begin{IEEEbiography}[{\includegraphics[width=1in,height=1.25in,clip,keepaspectratio]{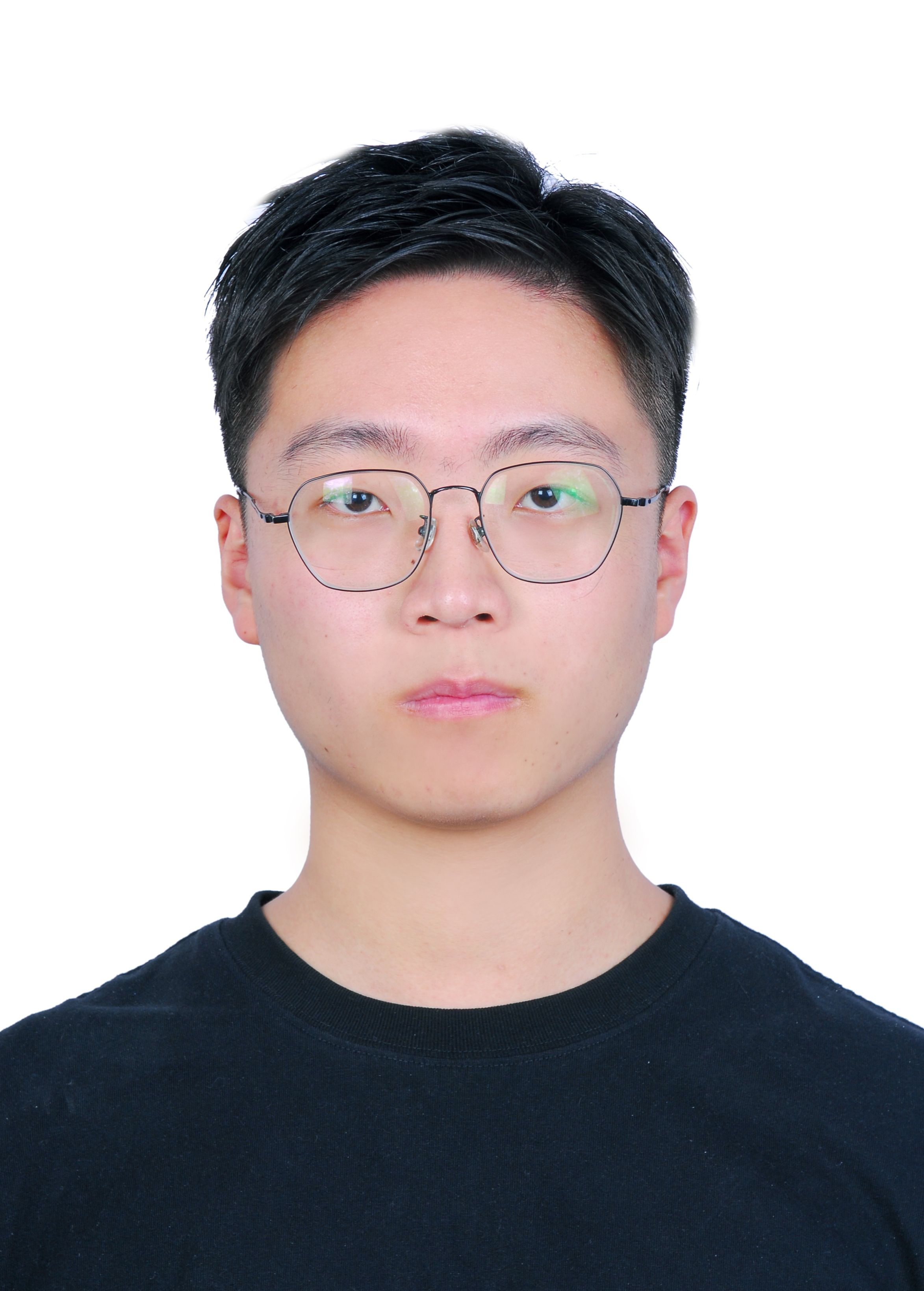}}]{Hongjie Liao} is currently a research assistant after master graduation at The Chinese University of Hong Kong, Shenzhen (CUHK-Shenzhen). He received his B.E. degree in 2022 from Beijing University of Posts and Communications and his Master's degree in 2024 from CUHK-Shenzhen. His research interests include 3D reconstruction and generation of objects and humans.
\end{IEEEbiography}
\vspace{-40pt}

\begin{IEEEbiography}[{\includegraphics[width=1in,height=1.25in,clip,keepaspectratio]{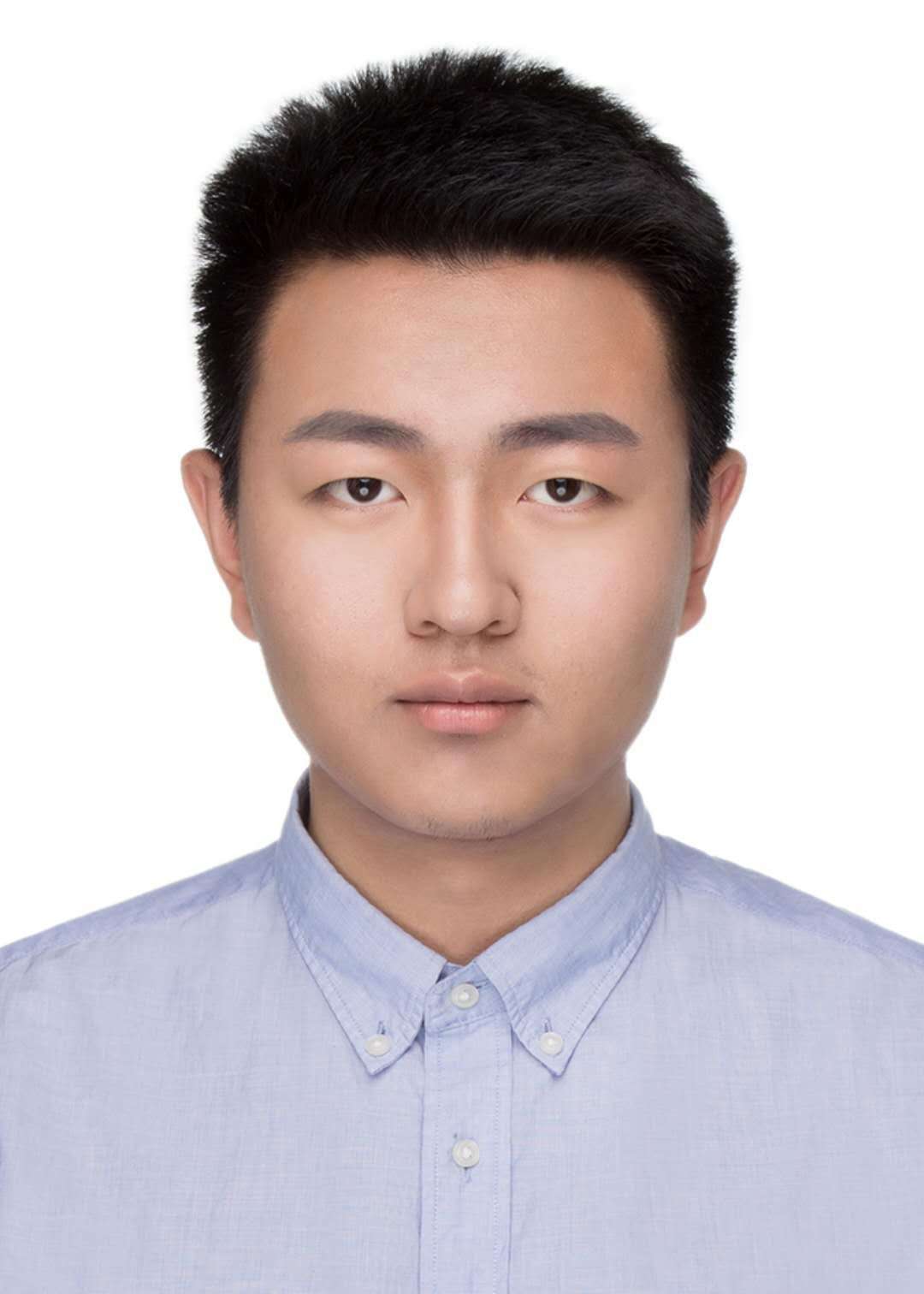}}]{Xihe Yang}
 received his Bachelor's degree from the University of Science and Technology of China in 2019 and Master degree from Rensselaer Polytechnic Institute, USA, in 2021. He is pursuing the PhD degree at The Chinese University of Hong Kong, Shenzhen. His research interests include 3D avatar reconstruction and generation.
\end{IEEEbiography}
\vspace{-40pt}

\begin{IEEEbiography}[{\includegraphics[width=1in,height=1.25in,clip,keepaspectratio]{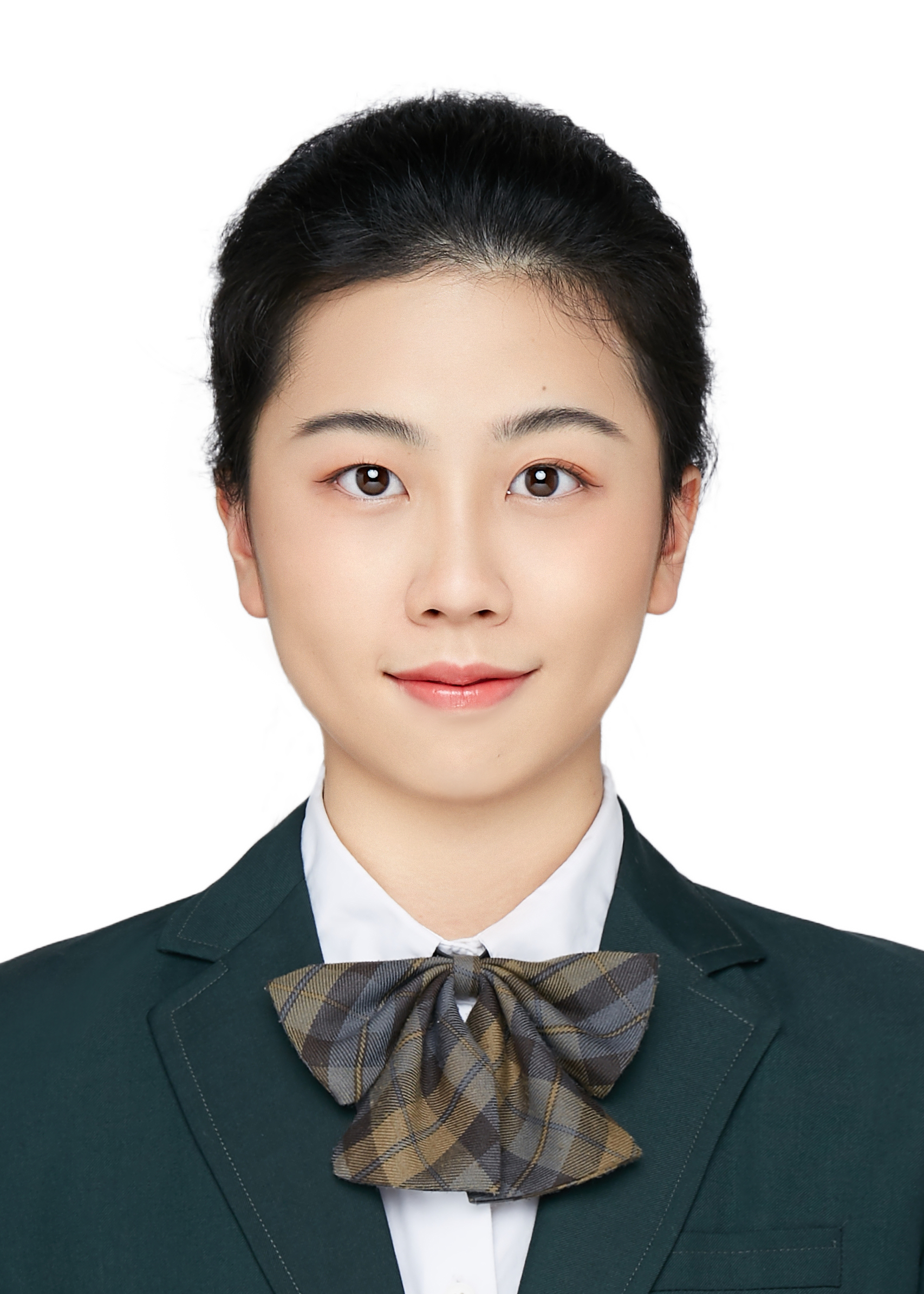}}]{Keru Zheng} received her Bachelor's degree from The Chinese University of Hong Kong, Shenzhen in 2025. She is currently working as a Research Assistant at The Chinese University of Hong Kong, Shenzhen. Her research interests include image generation, image editing, and medical image processing.
\end{IEEEbiography}
\vspace{-40pt}

\begin{IEEEbiography}[{\includegraphics[width=1in,height=1.25in,clip,keepaspectratio]{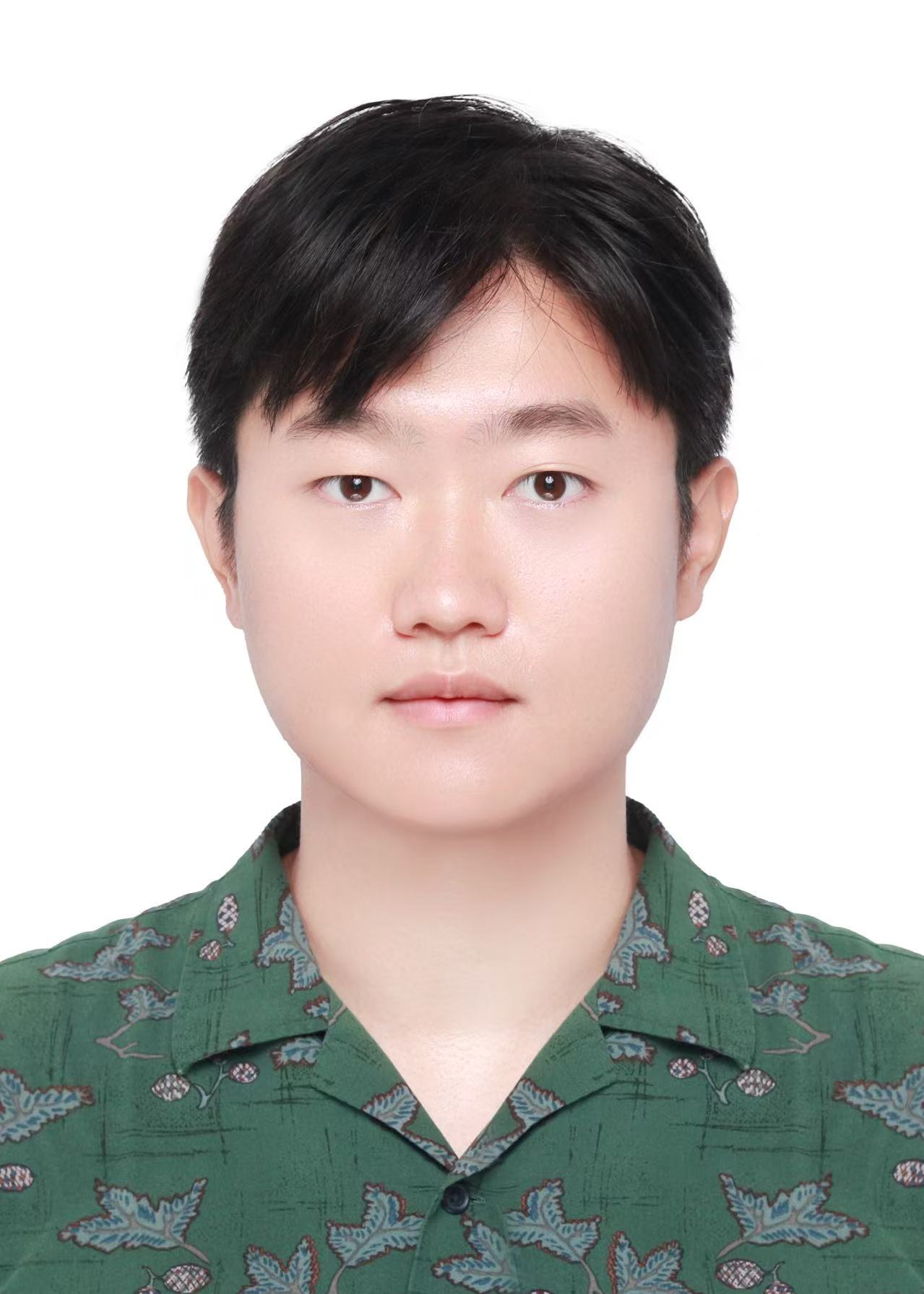}}]{Heyuan Li} received his Bachelor's degree from the University of Electronic Science and Technology of China in 2020. He obtained a Master of Computing degree from National University of Singapore in 2023. He is pursuing a PhD degree at The Chinese University of Hong Kong, Shenzhen and working on human-centric 3D computer vision and graphics.
\end{IEEEbiography}
\vspace{-40pt}

\begin{IEEEbiography}[{\includegraphics[width=1in,height=1.25in,clip,keepaspectratio]{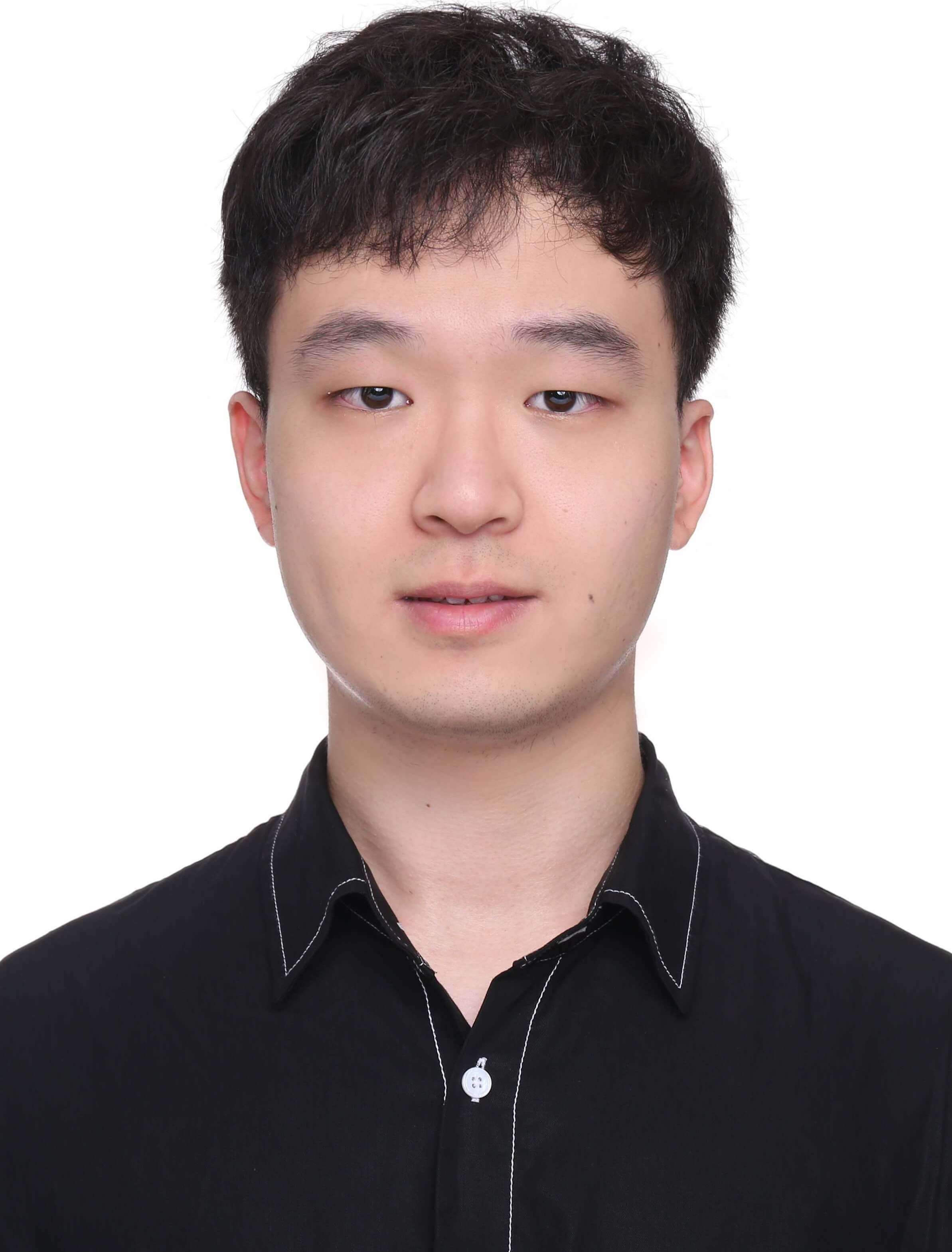}}]{Yihao Zhi} is currently pursuing his PhD degree at The Chinese University of Hong Kong, Shenzhen. He received his B.E. degree in 2020 from Shanghai University and Master degree in 2023 from ShanghaiTech University. His research interests include human reconstruction and animation.
\end{IEEEbiography}
\vspace{-40pt}

\begin{IEEEbiography}[{\includegraphics[width=1in,height=1.25in,clip,keepaspectratio]{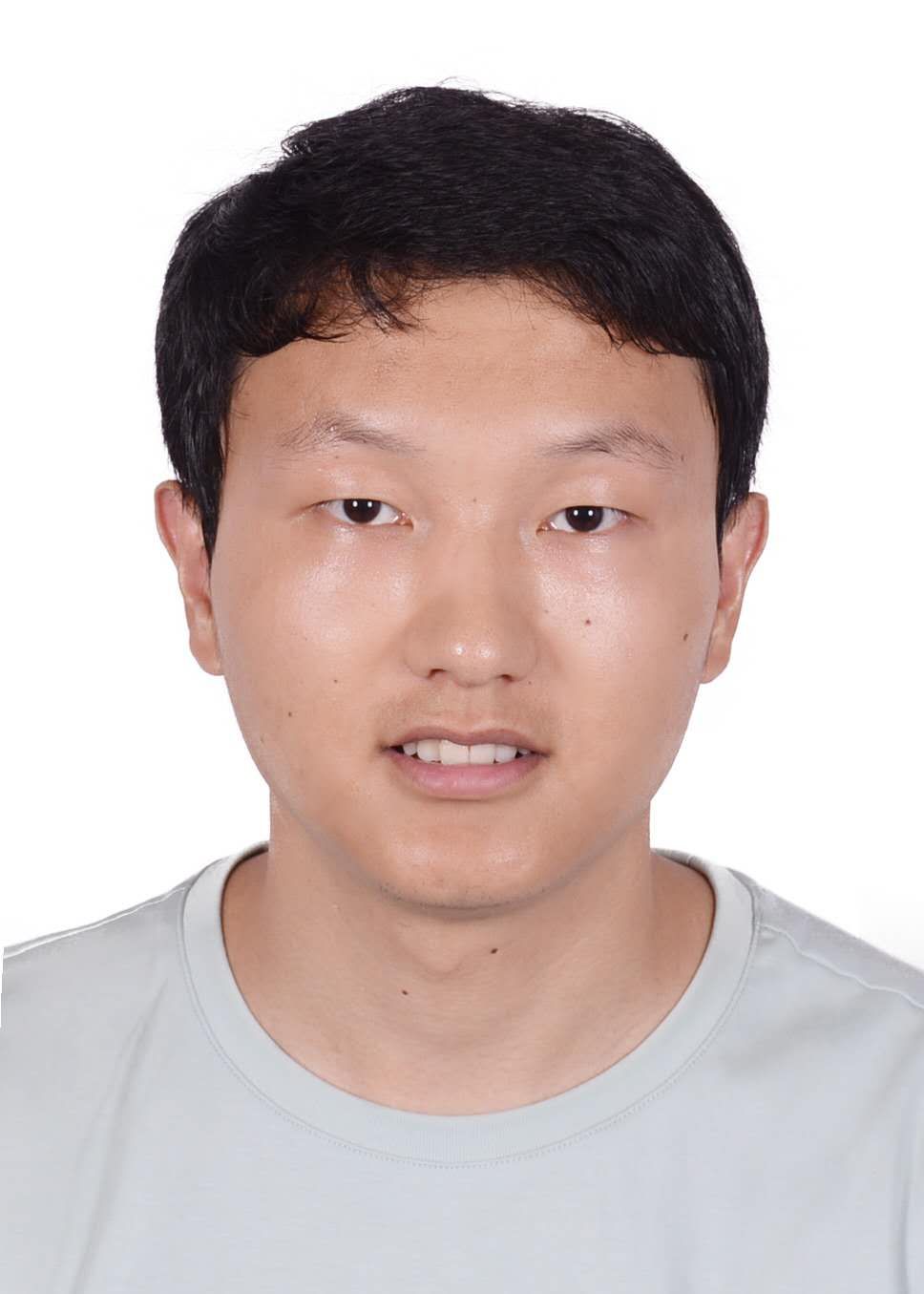}}]{Shuliang Ning} received his Bachelor’s degree and PhD degree from The Chinese University of Hong Kong, Shenzhen. He is currently a researcher at ByteDance. His research interests include image and video generation.
\end{IEEEbiography}
\vspace{-40pt}

\begin{IEEEbiography}[{\includegraphics[width=1in,height=1.25in,clip,keepaspectratio]{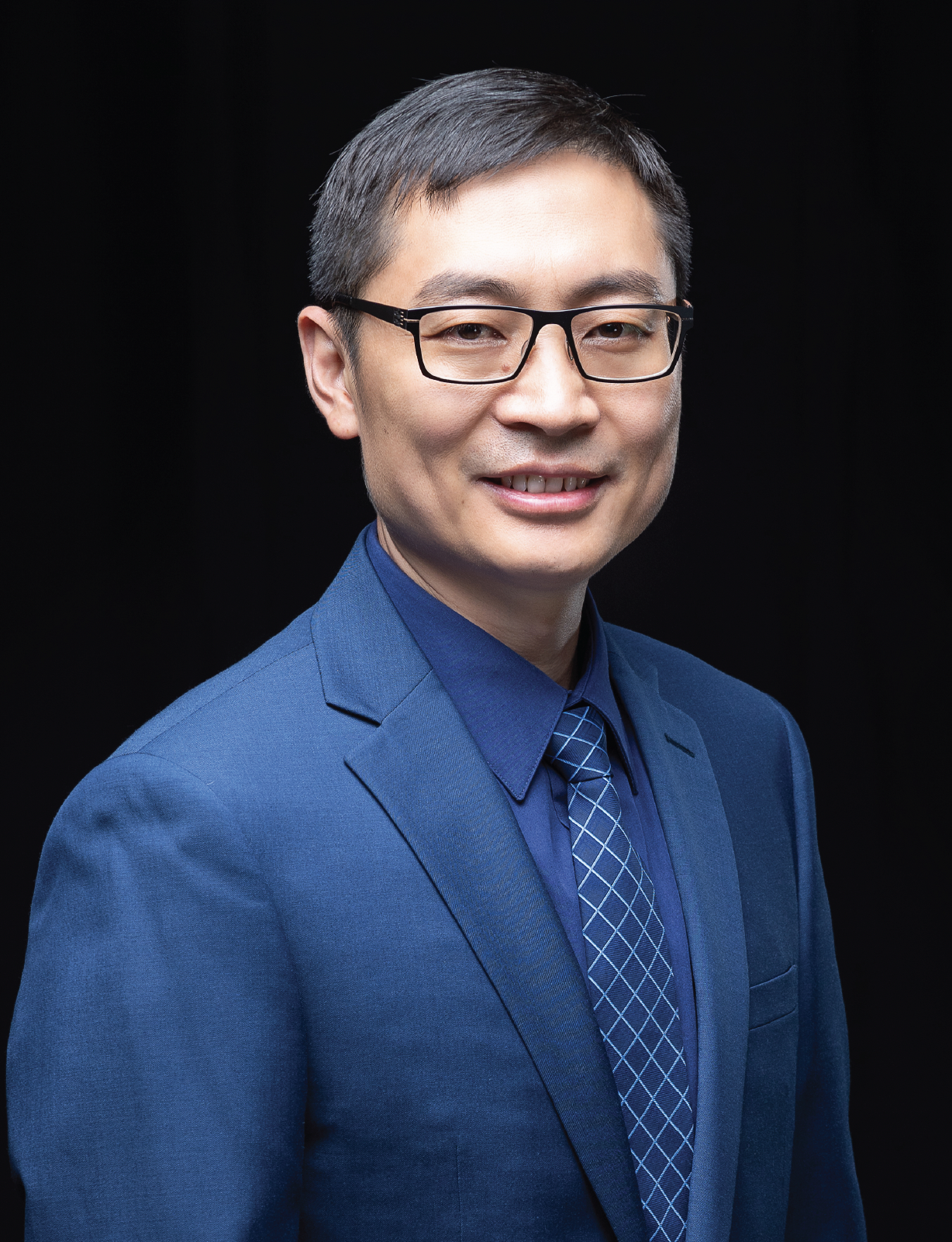}}]{Shuguang Cui} (Fellow, IEEE) received his Ph.D in Electrical Engineering from Stanford University, California, USA, in 2005. Afterwards, he has been working as assistant, associate, full, Chair Professor in Electrical and Computer Engineering at the Univ. of Arizona, Texas A\&M University, UC Davis, and CUHK at Shenzhen respectively. He has also served as the Executive Dean for the School of Science and Engineering and is currently the Director for Future Network of Intelligence Institute (FNii) at CUHK, Shenzhen, and the Executive Vice Director at Shenzhen Research Institute of Big Data. His current research interests focus on data driven large-scale system control and resource management, large data set analysis, IoT system design, energy harvesting based communication system design, and cognitive network optimization. He was selected as the Thomson Reuters Highly Cited Researcher and listed in the Worlds’ Most Influential Scientific Minds by ScienceWatch in 2014. He was the recipient of the IEEE Signal Processing Society 2012 Best Paper Award. He has served as the general co-chair and TPC co-chairs for many IEEE conferences. He has also been serving as the area editor for IEEE Signal Processing Magazine, and asso-ciate editors for IEEE Transactions on Big Data, IEEE Transactions on Signal Processing, IEEE JSAC Series on Green Communications and Networking, and IEEE Transactions on Wireless Communications. He has been the elected member for IEEE Signal Processing Society SPCOM Technical Committee (2009 2014) and the elected Chair for IEEE ComSoc Wireless Technical Committee (2017 2018). He is a member of the Steering Committee for IEEE Transactions on Big Data and the Chair of the Steering Committee for IEEE Transactions on Cognitive Communications and Networking. He was also a member of the IEEE ComSoc Emerging Technology Committee. He was elected as an IEEE Fellow in 2013, an IEEE ComSoc Distinguished Lecturer in 2014, and IEEE VT Society Distinguished Lecturer in 2019. In 2020, he won the IEEE ICC best paper award, ICIP best paper finalist, the IEEE Globecom best paper award. In 2021, he won the IEEE WCNC best paper award.
\end{IEEEbiography}
\vspace{-40pt}

\begin{IEEEbiography}[{\includegraphics[width=1in,height=1.25in,clip,keepaspectratio]{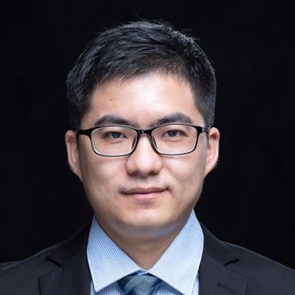}}]{Xiaoguang Han} (Member, IEEE) is now an Assistant Professor at The Chinese University of Hong Kong, Shenzhen. He received his PhD degree from The University of Hong Kong. His research interests cover both computer graphics and computer vision. He has published over 100 top tier papers. He serves as Associate Editors for IEEE TVCG and Computer $\&$ Graphics. 
\end{IEEEbiography}

\end{document}